\documentclass{easychair}

\usepackage{doc}

\usepackage{pifont}
\usepackage{xcolor}
\definecolor{second}{HTML}{007000}

\usepackage{comment}

\usepackage{amssymb}
\usepackage{pifont}
\usepackage{booktabs}
\usepackage{longtable}
\usepackage{pgfplotstable}
\usepackage{titlesec}
\usepackage{multirow}
\usepackage{subcaption}
\usepackage{fancyvrb}
\usepackage{cleveref}
\usepackage{threeparttable}
\usepackage{wrapfig}
\usepackage{makecell}
\usepackage{hyperref}
\usepackage{tablefootnote}

\pgfplotsset{compat=1.16}

\newcommand{\xmark}{\ding{55}}

\pgfkeysifdefined{/pgfplots/table/output empty row/.@cmd}{
    \pgfplotstableset{
        empty header/.style={
          every head row/.style={output empty row},
        }
    }
}{
    \pgfplotstableset{
        empty header/.style={
            typeset cell/.append code={%
                \ifnum\pgfplotstablerow=-1 %
                    \pgfkeyssetvalue{/pgfplots/table/@cell content}{}%
                \fi
            }
        }
    }
}

\titlespacing*{\paragraph}
{0pt}{2pt}{5pt}

\title{The Fourth International Verification of Neural Networks Competition (VNN-COMP 2023): Summary and Results}%

\author{Christopher Brix\inst{1}
        \and Stanley Bak\inst{2}
        \and Changliu Liu\inst{3}
        \and Taylor T. Johnson\inst{4}
}

\institute{
    RWTH Aachen University, Aachen, Germany\\
    \email{brix@cs.rwth-aachen.de}
    \and
    Stony Brook University, Stony Brook, New York, USA\\
    \email{stanley.bak@stonybrook.edu}
    \and
    Carnegie Mellon University, Pittsburgh, Pennsylvania, USA\\
    \email{cliu6@andrew.cmu.edu}
    \and
    Vanderbilt University, Nashville, Tennessee, USA\\
    \email{taylor.johnson@vanderbilt.edu}
 }

\authorrunning{C. Brix, S. Bak, C. Liu, T. Johnson}
\titlerunning{VNN-COMP 2023 Report}

\date{\phantom{date}}

\begin{document}

\maketitle

\begin{abstract}
This report summarizes the 4th International Verification of Neural Networks Competition (VNN-COMP 2023), held as a part of the 6th Workshop on Formal Methods for ML-Enabled Autonomous Systems (FoMLAS), which was collocated with the 35th International Conference on Computer-Aided Verification (CAV). 
The VNN-COMP is held annually to facilitate the fair and objective comparison of state-of-the-art neural network verification tools, encourage the standardization of tool interfaces, and bring together the neural network verification community.
To this end, standardized formats for networks (ONNX) and specification (VNN-LIB) were defined, tools were evaluated on equal-cost hardware (using an automatic evaluation pipeline based on AWS instances), and tool parameters were chosen by the participants before the final test sets were made public.
In the 2023 iteration, 7 teams participated on a diverse set of 10 scored and 4 unscored benchmarks.
This report summarizes the rules, benchmarks, participating tools, results, and lessons learned from this iteration of this competition.
\end{abstract}



\section{Introduction}
\label{sec:introduction}


Deep learning based systems are increasingly being deployed in a wide range of domains, including recommendation systems, computer vision, and autonomous driving. While the nominal performance of these methods has increased significantly over the last years, often even surpassing human performance, they largely lack formal guarantees on their behavior. 
However, in safety-critical applications, including autonomous systems, robotics, cybersecurity, and cyber-physical systems, such guarantees are essential for certification and confident deployment. 

While the literature on the verification of traditionally designed systems is wide and successful, neural network verification remains an open problem, despite significant efforts over the last years. In 2020, the International Verification of Neural Networks Competition (VNN-COMP) was established to facilitate comparison between existing approaches, bring researchers working on this problem together, and help shape future directions of the field. In 2023, the 4th iteration of the annual VNN-COMP\footnote{\url{https://sites.google.com/view/vnn2023/home}} was held as a part of the 6th Workshop on Formal Methods for ML-Enabled Autonomous Systems (FoMLAS) that was collocated with the 35th International Conference on Computer-Aided Verification (CAV). 

This fourth iteration of the VNN-COMP continues last year's trend of increasing standardization and automatization, aiming to enable a fair comparison between the participating tools and to simplify the evaluation of a large number of tools on a variety of (real-world) problems. 
As in the last iteration, VNN-COMP 2023 standardizes 1) neural network and specification formats, ONNX for neural networks and VNN-LIB \cite{vnnlib} for specifications, 2) evaluation hardware, providing participants the choice of a range of cost-equivalent AWS instances with different trade-offs between CPU and GPU performance, and 3) evaluation pipelines, enforcing a uniform interface for the installation and evaluation of tools. 

The competition was kicked off with the solicitation for participation in February 2023. By March, several teams had registered, allowing the rule discussion to be finalized in April 2023 (see an overview in \Cref{sec:rules}). From April to June 2023, benchmarks were proposed and discussed. Meanwhile, the organizing team decided to continue using AWS as the evaluation platform and started to implement an automated submission and testing system for both benchmarks and tools. By mid-July 2023, seven teams submitted their tools and the organizers evaluated all entrants to obtain the final results, discussed in \Cref{sec:results} and presented at FoMLAS on July 18, 2023. 
Discussions were structured into three issues on the official GitHub repository\footnote{\url{https://github.com/stanleybak/vnncomp2023/issues}}: rules discussion, benchmarks discussion, and tool submission. All submitted benchmarks\footnote{\url{https://github.com/ChristopherBrix/vnncomp2023_benchmarks}} and final results\footnote{\url{https://github.com/ChristopherBrix/vnncomp2023_results}} were aggregated in separate GitHub repositories. 

The remainder of this report is organized as follows: \Cref{sec:rules} discusses the competition rules, \Cref{sec:participants} lists all participating tools, \Cref{sec:benchmarks} lists all benchmarks,  \Cref{sec:results} summarizes the results, and \Cref{sec:conclusion} concludes the report, discussing potential future improvements.

\newpage

\section{Rules}
\label{sec:rules}

\paragraph*{Terminology}
An \emph{instance} is defined by a property specification (pre- and post-condition), a network, and a timeout). 
For example, one instance might consist of an MNIST classifier with one input image, a given local robustness threshold $\epsilon$, and a specific timeout.
A \emph{benchmark} is defined as a set of related instances.
For example, one benchmark might consist of a specific MNIST classifier with 100 input images, potentially different robustness thresholds $\epsilon$, and one timeout per input.

\paragraph*{Run-time caps}
Run-times are capped on a per-instance basis, i.e., any verification instance will timeout (and be terminated) after at most X seconds, determined by the benchmark proposer. These can be different for each instance. 
The total per-benchmark runtime (sum of all per-instance timeouts) may not exceed 6 hours per benchmark. 
For example, a benchmark proposal could have six instances with a one-hour timeout, or 100 instances with a 3.6-minute timeout, each.
To enable a fair comparison, we measure the startup overhead for each tool by running it on a range of tiny networks and subtract the minimal overhead from the total runtime.

\paragraph*{Hardware}
To allow for comparability of results, all tools were evaluated on equal-cost hardware using  Amazon Web Services (AWS).
Each team could decide between a range of AWS instance types (see \Cref{tab:instances}) providing a CPU, GPU, or mixed focus.


\begin{table}[h]
\centering
\caption{Available AWS instances.}\label{tab:instances}
\renewcommand{\arraystretch}{1.1}
\scalebox{0.985}{
\begin{tabular}{lccc} \toprule
         &vCPUs & RAM [GB] & GPU \\ 
         \midrule
         p3.2xlarge  & 8 & 61 & V100 GPU with 16 GB memory \\
         m5.16xlarge  & 64 & 256 & \xmark \\
         g5.8xlarge  & 32 & 128 & A10G GPU with 24 GB memory \\
         \bottomrule
\end{tabular}
}
\end{table}

\paragraph{Scoring} 
The final score is aggregate as the sum of all benchmark scores.
Each benchmark score is the number of points (sum of instance scores discussed below) achieved by a given tool, normalized by the maximum number of points achieved by any tool on that benchmark. 
Thus, the tool with the highest sum of instance scores for a benchmark will get a benchmark score of 100, ensuring that all benchmarks are weighted equally, regardless of the number of constituting instances.

\paragraph*{Instance score}
\label{sec:scoring}
Each instance is scored is as follows: 
\begin{itemize}\setlength{\itemsep}{0pt}
    \item Correct hold (property proven): 10 points;
    \item Correct violated (counterexample found): 10 points;
    \item Incorrect result: -150 points (penalty increased compared to 2022).
\end{itemize}
However, the ground truth for any given instance is generally not known a priori. In the case of disagreement between tools, we, therefore, place the burden of proof on the tool claiming that a specification is violated, i.e. that a counterexample can be found, and deem it correct exactly if it produces a valid counterexample.

%
The provided counterexamples were supposed to define both the input and the resulting output of the networks.
However, for some tools and instances, the output definition was either missing or differed from the network output as computed by the onnxruntime package used to evaluate counterexamples (by performing inference given the inputs).
The competition rules were ambiguous how this would be handled. We decided to discard all outputs in the counterexample files and base the evaluation solely on the given inputs and their respective outputs as computed by the onnxruntime.
A ranking using the alternative evaluation, where incorrect or missing outputs result in a penalty can be found in Appendix~\ref{sec:alternative_ranking}.

\paragraph*{Time bonus}
As opposed to previous years, no time bonus was awarded.
Instead, all tools that are compute the correct result within the time limit receive the same amount of points.

\paragraph{Overhead Correction}
The overhead of tools was measured, but only used to adapt the timeouts. It did not influence the scores, as no time bonus was awarded.
To measure the tool-specific overhead, we created trivial network instances and included those in the measurements. We then observed the minimum verification time over all instances and considered that to be the overhead time for the tool.

\paragraph{Format}
As in 2022, we standardized neural networks to be in \texttt{onnx} format, specifications in \texttt{vnnlib} format, and counterexamples in a format similar to the \texttt{vnnlib} format.
Further, tool authors were required to provide scripts fully automating the installation process of their tool, including the acquisition of any licenses that might be needed. Similar to the previous year, a preparation and execution script had to be provided for running their tool on a specific instance consisting of a network file, specification file, and timeout.
The specifications are interpreted as definitions of counterexamples, meaning that a property is proven ``correct'' if the specification is shown to be unsatisfiable, conversely, the property is shown to be violated if a counterexample fulfilling the specification is found. 
Specifications consisted of disjunctions over conjunctions in both pre- and post-conditions, allowing a wide range of properties from adversarial robustness over multiple hyper-boxes to safety constraints to be encoded.
For example, robustness with respect to inputs in a hyper-box had to be encoded as disjunctive property, where any of the other classes is predicted.

\newpage

\section{Participants}
\label{sec:participants}
We list the tools and teams that participated in the VNN-COMP 2023 in \Cref{tab:tools} and reproduce their own descriptions of their tools below.



\begin{table*}[h]
\begin{center}
\begin{minipage}{\linewidth}
\caption{Summary of the key features of participating tools. The hardware column describes the used AWS instance with \texttt{p3} and \texttt{g5} making GPUs available, see \Cref{tab:instances} for more details. Licenses refer to the external licenses required to use the corresponding tool, not the licensing of the tool itself.}%
\label{tab:tools}%
\renewcommand{\arraystretch}{1.25}
\resizebox{\textwidth}{!}{
\begin{tabular}{lcm{5.0cm}ccc} \toprule
         Tool & References & Organizations & Place & Hardware & Licenses\\
         \midrule
         \textbf{$\boldsymbol{\alpha}$,$\boldsymbol{\beta}$-CROWN} & \cite{xu2020automatic,xu2021fast,wang2021betacrown,zhang2022general,shi2023formal} & UIUC, CMU, UCLA, Drexel, Columbia, RWTH Aachen, Sun Yat-Sen, UMich. & \textbf{1} & \texttt{g5} & GUROBI\\
%
         FastBATLLNN & \cite{FerlezKS22} & University of California & 7 & t2 & -\\
         Marabou & \cite{KatzHIJLLSTWZDK19} & Hebrew University of Jerusalem, Stanford University, NRI Secure & 2 & m5 & GUROBI\\
         NeuralSAT & \cite{duong2023dpllt} & George Mason University & 4 & g5 & GUROBI\\
         nnenum & \cite{bak2020cav,bak2021nnenum} & Stony Brook University & 5 & m5 & - \\
         NNV & \cite{tran2020cav_tool,manzanas2023cav} & Vanderbilt University, University of Nebraska & 6 & m5 & MATLAB\\
         PyRAT &  \cite{pyrat-website} & Universite Paris-Saclay, CEA, List & 3 & m5 & - \\
         \bottomrule
\end{tabular}
}
\end{minipage}
\end{center}
\end{table*}


\subsection{$\alpha,\!\beta$-CROWN}
\paragraph*{Team} $\alpha,\!\beta$-CROWN is developed by a multi-institutional team from UIUC, CMU, UCLA, Drexel, Columbia University, RWTH Aachen University, Sun Yat-Sen University, and the University of Michigan.
\begin{itemize}
    \item Team leaders: Huan Zhang (UIUC/CMU) and Linyi Li (UIUC)
    \item VNN-COMP 2023 team members: Zhouxing Shi (UCLA), Christopher Brix (RWTH Aachen University), Kaidi Xu (Drexel University), Xiangru Zhong (Sun Yat-Sen University), Qirui Jin (University of Michigan), Zhuowen Yuan (UIUC).
\end{itemize}

\paragraph*{Description} $\alpha,\!\beta$-CROWN (\texttt{alpha-beta-CROWN}) is an efficient neural network verifier based on the linear bound propagation framework, built on a series of works on bound-propagation-based neural network verifiers:  CROWN~\cite{zhang2018efficient,xu2020automatic}, $\alpha$-CROWN~\cite{xu2021fast}, $\beta$-CROWN~\cite{wang2021betacrown}, GCP-CROWN~\cite{zhang2022general}, and nonlinear branch-and-bound~\cite{shi2023formal}.
The core techniques in $\alpha,\!\beta$-CROWN combine the efficient and GPU-accelerated linear bound propagation method with branch-and-bound methods. Previously, branch-and-bound was conducted for ReLU neural networks only. In this year, we extended branch-and-bound for general nonlinearities~\cite{shi2023formal}, and enabled strong verification for generic nonlinear computation graphs, such as in the ML4ACOPF benchmark.


The linear bound propagation algorithms in $\alpha,\!\beta$-CROWN are based on our \texttt{auto\_LiRPA} library~\cite{xu2020automatic}, which supports general neural network architectures (including convolutional layers, pooling layers, residual connections, recurrent neural networks, and Transformers) and a wide range of activation functions (e.g., ReLU, tanh, trigonometric functions, sigmoid, max pooling and average pooling), and is efficiently implemented on GPUs with Pytorch and CUDA. We jointly optimize intermediate layer bounds and final layer bounds using gradient ascent (referred to as $\alpha$-CROWN or optimized CROWN/LiRPA~\cite{xu2021fast}). Most importantly, we use branch and bound~\cite{bunelunified2018} (BaB) and incorporate split constraints in BaB into the bound propagation procedure efficiently via the $\beta$-CROWN algorithm~\cite{wang2021betacrown}, use cutting-plane method in GCP-CROWN~\cite{zhang2022general} to further tighten the bound, and support general nonlinearities in the branch-and-bound~\cite{shi2023formal}.
For smaller networks, we also use a mixed integer programming (MIP) formulation~\cite{Tjeng2019EvaluatingRO} combined with tight intermediate layer bounds from $\alpha$-CROWN (referred to as $\alpha$-CROWN + MIP~\cite{zhang2022general}). The combination of efficient, optimizable and GPU-accelerated bound propagation with BaB produces a powerful and scalable neural network verifier.

\paragraph*{Link} \url{https://github.com/Verified-Intelligence/alpha-beta-CROWN} (latest version)
\paragraph*{Competition submission} \url{https://github.com/Verified-Intelligence/alpha-beta-CROWN_vnncomp23} (only for reproducing competition results; please use the latest version for other purposes)
\paragraph*{Hardware and licenses} CPU and GPU with 32-bit or 64-bit floating point; Gurobi license required for the \texttt{gtrsb} benchmark.
\paragraph*{Participated benchmarks} All benchmarks.

\subsection{FastBATLLNN}
\paragraph{Team} James Ferlez (Developer), Haitham Khedr (Tester), and Yasser Shoukry (Supervisor) (University of California, Irvine)
\paragraph{Description} FastBATLLNN \cite{FerlezKS22} is a fast verifier of box-like (hyper-rectangle) output properties for Two-Level Lattice (TLL) Neural Networks (NN). FastBATLLNN uses both the unique semantics of the TLL architecture and the decoupled nature of box-like output constraints to provide a fast, polynomial-time verification algorithm: that is, polynomial-time in the number of neurons in the TLL NN to be verified (for a fixed input dimension). FastBATLLNN fundamentally works by converting the TLL verification problem into a region enumeration problem for a hyperplane arrangement (the arrangement is jointly derived from the TLL NN and the verification property). However, as FastBATLLNN leverages the unique properties of TLL NNs and box-like output constraints, its use is necessarily limited to verification problems formulated in those terms. Hence, FastBATLLNN can only compete on the \texttt{tllverifybench} benchmark.
\paragraph{Link} \url{https://github.com/jferlez/FastBATLLNN-VNNCOMP}
\paragraph*{Commit}
6100258b50a3fadf9792aec4ccf39ed14778e338
\paragraph{Hardware and licenses} CPU; no licenses required
\paragraph{Participated benchmarks} \texttt{tllverifybench}

\subsection{Marabou}
\paragraph{Team} Haoze Wu (Stanford University), Clark Barrett (Stanford University), Guy Katz (Hebrew University of Jerusalem)
\paragraph{Description}  Marabou~\cite{katz2019marabou} is a user-friendly Neural Network Verification toolkit that can answer queries about a network’s properties by encoding and solving these queries as constraint satisfaction problems. It has both Python/C++ APIs through which users can load neural networks and define arbitrary linear properties over the neural network. Marabou supports many different linear, piecewise-linear, and non-linear~\cite{wu2022toward,wei2023convex} operations and architectures (e.g., FFNNs, CNNs, residual connections, Graph Neural Networks~\cite{vegas}). 

Under the hood, Marabou employs a uniform solving strategy for a given verification query. In particular, Marabou performs complete analysis that employs a specialized convex optimization procedure~\cite{wu2022efficient} and abstract interpretation~\cite{DeepPoly:19,vegas}. It also uses the Split-and-Conquer algorithm~\cite{wu2020parallelization} for parallelization.
\footnote{Thanks to the authors of the $\alpha-\beta$-CROWN team, an unsoundness issue of the competition version of Marabou on the \texttt{ViT} benchmarks was discovered. The networks in that benchmark contain bilinear and softmax connections. For this benchmark, the competition version of Marabou first performs DeepPoly-style abstract interpretation and then encodes the verification problem in the Gurobi optimizer. It turns out that Gurobi can report ``Infeasible'' on benchmarks where counter-examples are expected. The Marabou team is actively looking into resolving this issue. }

\paragraph{Link} \url{https://github.com/NeuralNetworkVerification/Marabou}
\paragraph*{Commit} 1a3ca6010b51bba792ef8ddd5e1ccf9119121bd8
\paragraph{Hardware and Licenses} CPU, no license required. Can also be accelerated with Gurobi (which requires a license)
\paragraph{Participated benchmarks} \texttt{acasxu}, \texttt{cgan}, \texttt{collins\_rul\_cnn}, \texttt{dist\_shift}, \texttt{ml4acopf}, \texttt{nn4sys}, \texttt{tllverifybench}, \texttt{traffic\_signs\_recognition}, \texttt{vggnet16}, \texttt{vit}.

\subsection{nnenum}
\paragraph*{Team} Ali Arjomandbigdeli (Student), Stanley Bak (Supervisor) (Stony Brook University)

\paragraph*{Description} 
The nnenum tool~\cite{bak2021nnenum} uses multiple levels of abstraction to achieve high-performance verification of ReLU networks without sacrificing completeness~\cite{bak2020vnn}. 
The core verification method is based on reachability analysis using star sets~\cite{tran2019fm}, combined with the ImageStar method~\cite{tran2020cav} to propagate stes through all linear layers supported by the ONNX runtime, such as convolutional layers with arbitrary parameters.  
The tool is written in Python 3 and uses GLPK for LP solving.
New this year, we added support for single lower and single upper bounds propagation in addition to zonotopes, similar to the DeepPoly method or the CROWN approach.
We also added an option to use Gurobi instead of GLPK for LP solving.

\paragraph*{Link}
\url{https://github.com/aliabigdeli/nnenum}

\paragraph*{Commit}
2e14cdf09f66d8b3c4622ad29d8bedd815d056e8

\paragraph*{Hardware and licences}
CPU, Gurobi license (optional)

\paragraph*{Participated benchmarks}
\texttt{acasxu}, \texttt{cgan}, \texttt{collins-rul-cnn}, \texttt{nn4sys}, \texttt{tllverifybench}, \texttt{vggnet16}.

\subsection{NNV}
\paragraph*{Team} Diego Manzanas Lopez (Vanderbilt University), Neelanjana Pal (Vanderbilt University), Samuel Sasaki (Vanderbilt University), Hoang-Dung Tran (University of Nebraska-Lincoln), Taylor T. Johnson (Vanderbilt University)
\paragraph*{Description} The Neural Network Verification (NNV) Tool~\cite{tran2020cav_tool,manzanas2023cav} is a formal verification software tool for deep learning models and cyber-physical systems with neural network components written in MATLAB and available at \url{https://github.com/verivital/nnv}. NNV uses a star-set state-space representation and reachability algorithm that allows for a layer-by-layer computation of exact or overapproximate reachable sets for feed-forward~\cite{tran2019fm}, convolutional~\cite{tran2020cav}, semantic segmentation (SSNN)~\cite{tran2021cav}, and recurrent (RNN)\cite{tran2023hscc} neural networks, as well as neural network control systems (NNCS)~\cite{tran2019emsoft,tran2020cav_tool} and neural ordinary differential equations (Neural ODEs)~\cite{manzanas2022formats}. 
The star-set based algorithm is naturally parallelizable, which allows NNV to be designed to perform efficiently on multi-core platforms. Additionally, if a particular safety property is violated, NNV can be used to construct and visualize the complete set of counterexample inputs for a neural network (exact-analysis). 
%
For this competition, we make use of the same verification approach across all properties (i.e., no fine-tuning for individual benchmarks). First, we perform a simulation-guided search for counterexamples for a fixed number of samples. If no counterexamples are found (i.e., demonstrate that the property is SAT), then we utilize an iterative refinement approach using reachability analysis to verify the property (UNSAT). this consists of performing reachability analysis using a relax-approximation method~\cite{tran2021cav}, if not verified, then a less conservative approximation based on zonotope pre-filtering approach~\cite{tran2021fac}, and finally using the exact analysis when possible~\cite{tran2020cav} until the specification is verified or there is a timeout. 

\paragraph*{Link} \url{https://github.com/verivital/nnv}

\paragraph*{Commit} fb858b070d7c2fb1f036ac7fc374e1b9dfb5055e

\paragraph*{Hardware and licenses} CPU, MATLAB license.

\paragraph*{Participated Benchmarks} 
\texttt{acasxu}, \texttt{cgan}, \texttt{collins-rul-cnn}, \texttt{dist-shift}, \texttt{nn4sys},\\ \texttt{tllverifybench}.

\subsection{NeuralSAT}
\paragraph*{Team} Hai Duong and Thanhvu Nguyen (George Mason).

\begin{wrapfigure}{r}{0.25\textwidth}
  \centering
  \vspace{-0.1in}
  \includegraphics[width=1\linewidth]{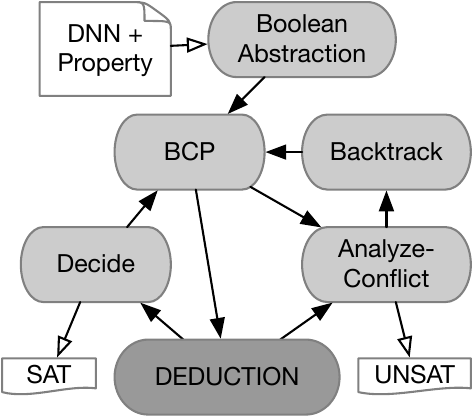}
  \vspace{-0.1in}  
\end{wrapfigure}
\paragraph*{Description} 
NeuralSAT~\cite{duong2023dpllt} integrates \emph{clause learning} in modern constraint solving to enhance performance and address SAT challenges like backtracking, with an \emph{abstraction-based theory solver} in SMT solving and abstraction-based DNN verification to expedite infeasibility checking.
%
The figure on the right gives an overview of NeuralSAT, which implements the DPLL(T) framework used in modern SMT solvers such as  Z3 and CVC. 
The design of NeuralSAT is inspired by the core algorithms used in SMT solvers such as CDCL components (light shades) and theory solving (dark shade). 
NeuralSAT can be considered as a native satisfiability solver for DNNs and inherits the benefits of modern SMT solvers.
The tool is written in Python and uses Gurobi for LP solving. 
Since this VNN-COMP participation, NeuralSAT has been updated with parallel DPLL search and many other optimizations.

\paragraph*{Link}
\url{https://github.com/dynaroars/neuralsat}

\paragraph*{Commit}
5693c3da130942283744ce56c2e74ac6c16eef94

\paragraph*{Hardware and licences}
GPU, Gurobi License

\paragraph*{Participated benchmarks}
\texttt{acasxu}, 
\texttt{cgan}, 
\texttt{collins-rul-cnn}, 
\texttt{dist-shift}, 
\texttt{nn4sys}, 
\texttt{vggnet16},
\texttt{tllverifybench}, 
\texttt{trafic-signs-recognition},
\texttt{reach-prob-density},
\texttt{metaroom}.

\subsection{PyRAT}

\paragraph*{Team} Augustin Lemesle, Julien Lehmann, Serge Durand, Zakaria Chihani (CEA-List)
\paragraph*{Description} 
PyRAT (Python Reachability Assessment Tool) is a tool for verifying various types of neural networks based on Python and abstract interpretation techniques. PyRAT can leverage multiple abstract domains such as Intervals, Zonotopes, or Polyhedras to efficiently compute bounds for different architectures of neural networks such as dense, convolutional, residual or recurrent neural networks. It supports ReLU, Sigmoid, Tanh, Softmax activation functions. PyRAT can efficiently work on both CPU and GPU and is sound w.r.t. floating points computation when using CPU. PyRAT is correct as the output bounds reached will always be a sound over approximation of the results.
Depending on the benchmark, several verification modes and domains can be selected. For smaller networks and problems, PyRAT can leverage branch and bound strategies on the inputs with heuristics like ReCIPH~\cite{durand2022reciph}. While on larger network, PyRAT will use GPU computation as well as local counterexample checking to verify or falsify a property.

\paragraph*{Link} \href{https://git.frama-c.com/pub/pyrat/-/tree/VNN_COMP2023}{https://git.frama-c.com/pub/pyrat}
\paragraph*{Commit} 50288df457653d8767c2d6f5481cc3e72e3d6727 
\paragraph*{Hardware and licenses} CPU and GPU, closed source CEA licence.
\paragraph*{Participated benchmarks} 
\texttt{acasxu}, 
\texttt{cgan}, 
\texttt{collins\_rul\_cnn}, 
\texttt{dist\_shift}, 
\texttt{nn4sys}, \\
\texttt{tllverifybench}, 
\texttt{traffic\_signs\_recognition}, 
\texttt{vggnet16}.

\newpage
\section{Benchmarks}
\label{sec:benchmarks}

In this section, we provide an overview of all benchmarks, reproducing the benchmark proposers' descriptions.

               
\begin{table}[h]
    \centering
    \caption{Overview of all scored benchmarks. }
    \label{tab:my_label}
    \resizebox{\textwidth}{!}{
    \renewcommand{\arraystretch}{1.4}
    \begin{tabular}{cccccc}
    \toprule
    Category &
    Benchmark &
    Application &
    Network Types &
    \# Params &
    Effective Input Dim
    \\
    \midrule
    \multirow{4.5}{*}{Complex} 
    & cGAN & \makecell{Image Generation \\ \& Image Prediction} & Complex (Conv. + Vision Transformer) & 500k - 68M & 5 \\
    & NN4Sys & \makecell{Dataset Indexing \\ \&  Cardinality Prediction}   & Complex (ReLU + Sigmoid)  & 33k - 37M & 1-308 \\
    & ml4acopf & Power System & Complex (ReLU + Trigonometric + Sigmoid) & 4k-680k & 22 - 402 \\
    & ViT & Vision & Conv. + Residual + Softmax + BatchNorm & 68k - 76k & 3072 \\
    \cmidrule(lr){1-6}
    \multirow{3}{*}{\makecell{CNN \\ \& ResNet}} 
    & Collins RUL CNN & Condition Based Maintenance & Conv. + ReLU, Dropout  & 60k - 262k   & 400 - 800 \\
    & VGGNet16 & Image Classification & Conv. + ReLU + MaxPool    & 138M & 150k \\
    & Traffic Signs Recognition & Image Classification & Conv. + Sign + MakPool + BatchNorm & 905k - 1.7M & 2.7k - 12k \\
    \cmidrule(lr){1-6} %
    \multirow{3.5}{*}{\makecell{FC}}
    & TLL Verify Bench & Two-Level Lattice NN & \makecell{Two-Level Lattice NN \\(FC. + ReLU)}  & 17k - 67M & 2 \\
    & Acas XU & Collision Detection & FC. + ReLU & 13k & 5 \\
    & Dist Shift & Distribution Shift Detection & FC. + ReLU + Sigmoid & 342k - 855k & 792 \\
    \bottomrule
    \end{tabular}
    }
\end{table}

\subsection{cGAN}
\paragraph*{Proposed by} Feiyang Cai, Ali Arjomandbigdeli, Stanley Bak (Stony Brook University)
\paragraph*{Motivation}
While existing neural network verification benchmarks focus on discriminative models, the exploration of practical and widely used generative networks remains neglected in terms of robustness assessment.
This benchmark introduces a set of image generation networks specifically designed for verifying the robustness of the generative networks.
\paragraph*{Networks}
The generative networks are trained using conditional generative adversarial networks (cGAN), whose objective is to generate camera images that contain a vehicle obstacle located at a specific distance in front of the ego vehicle, where the distance is controlled by the input distance condition.
The network to be verified is the concatenation of a generator and a discriminator.  The generator takes two inputs: 1) a distance condition (1D scalar) and 2) a noise vector controlling the environment (4D vector). The output of the generator is the generated image. The discriminator takes the generated image as input and outputs two values: 1) a real/fake score (1D scalar) and 2) a predicted distance (1D scalar).
Several different models with varying architectures (CNN and vision transformer) and image sizes (32x32, 64x64) are provided for different difficulty levels.
\paragraph*{Specifications}
The verification task is to check whether the generated image aligns with the input distance condition, or in other words, verify whether the input distance condition matches the predicted distance of the generated image.
In each specification, the inputs (condition distance and latent variables) are constrained in small ranges, and the output is the predicted distance with the same center as the condition distance but with slightly larger range.
\paragraph*{Link} \url{https://github.com/feiyang-cai/cgan_benchmark2023}

\pagebreak
\subsection{NN4Sys}
\paragraph*{Proposed by} the $\alpha,\!\beta$-CROWN team with collaborations with Cheng Tan and Haoyu He at Northeastern University.
\paragraph*{Application}
The benchmark contains networks for database learned index and learned cardinality
estimation, which maps input from various dimensions to a single scalar as output. 

\begin{itemize}

\item \textit{Background}: learned index and learned cardinality are all instances in neural networks for computer systems (NN4Sys), which are neural network based methods performing system operations. These classes of methods show great potential but have one drawback---the outputs of an NN4Sys model (a neural network) can be arbitrary, which may lead to unexpected issues in systems.

\item \textit{What to verify}: our benchmark provides multiple pairs of (1) trained NN4Sys model
and (2) corresponding specifications. We design these pairs with different parameters such
that they cover a variety of user needs and have varied difficulties for verifiers. 
We describe benchmark details in our NN4SysBench report~\cite{he2022Characterizing}: \url{http://naizhengtan.github.io/doc/papers/characterizing22haoyu.pdf}.

\item \textit{Translating NN4Sys applications to a VNN benchmark}: 
the original NN4Sys applications have sophisticated features that are hard to express.
We tailored the neural networks and their specifications to be suitable for VNN-COMP.
For example, learned index~\cite{kraska18case} contains multiple NNs in a tree structure that together serve one purpose.
However, this cascading structure is inconvenient/unsupported to verify
because there is a ``switch" operation---choosing one NN in the second stage
based on the prediction of the first stage's NN.
To convert learned indexes to a standard form, we merge the NNs into one larger NN.

\item \textit{A note on broader impact}: using NNs for systems is a broad topic, but many existing works
lack strict safety guarantees. We believe that NN Verification can help system developers gain confidence
to apply NNs to critical systems. We hope our benchmark can be an early step toward this vision.

\end{itemize}

\paragraph*{Networks}
This benchmark has six networks with different parameters: two for learned indexes
and four for learned cardinality estimation.
The learned index uses fully-connected feed-forward neural networks.
The cardinality estimation has a relatively sophisticated internal structure;
please see our NN4SysBench report (URL listed above) for details.

\paragraph*{Specifications}
For learned indexes,
the specification aims to check if the prediction error is bounded.
The specification is a collection of pairs of input and output intervals such that
any input in the input interval should be mapped to the corresponding output interval.
For learned cardinality estimation,
the specifications check the prediction error bounds (similar to the learned indexes)
and monotonicity of the networks.
By monotonicity specifications, we mean that for two inputs, the network should produce a larger
output for the larger input, which is required by cardinality estimation.

\paragraph{Link:} \url{https://github.com/Cli212/VNNComp22_NN4Sys}

\pagebreak





\subsection{Collins-RUL-CNN}
\paragraph*{Proposed by} Collins Aerospace, Applied Research \& Technology (\href{https://www.collinsaerospace.com/what-we-do/capabilities/technology-and-innovation/applied-research-and-technology}{website}).

\paragraph*{Motivation} Machine Learning (ML) is a disruptive technology for the aviation industry. This particularly concerns safety-critical aircraft functions, where high-assurance design and verification methods have to be used in order to obtain approval from certification authorities for the new ML-based products. Assessment of correctness and robustness of trained models, such as neural networks, is a crucial step for demonstrating the absence of unintended functionalities~\cite{ForMuLA, kirov2023formal}. The key motivation for providing this benchmark is to strengthen the interaction between the VNN community and the aerospace industry by providing a realistic use case for neural networks in future avionics systems~\cite{kirov2023benchmark}.

\paragraph*{Application} Remaining Useful Life (RUL) is a widely used metric in Prognostics and Health Management (PHM) that manifests the remaining lifetime of a component (e.g., mechanical bearing, hydraulic pump, aircraft engine). RUL is used for Condition-Based Maintenance (CBM) to support aircraft maintenance and flight preparation. It contributes to such tasks as augmented manual inspection of components and scheduling of maintenance cycles for components, such as repair or replacement, thus moving from preventive maintenance to \emph{predictive} maintenance (do maintenance only when needed, based on component’s current condition and estimated future condition). This could allow to eliminate or extend service operations and inspection periods, optimize component servicing (e.g., lubricant replacement), generate inspection and maintenance schedules, and obtain significant cost savings. Finally, RUL function can also be used in airborne (in-flight) applications to dynamically inform pilots on the health state of aircraft components during flight. Multivariate time series data is often used as RUL function input, for example, measurements from a set of sensors monitoring the component state, taken at several subsequent time steps (within a time window). Additional inputs may include information about the current flight phase, mission, and environment. Such highly multi-dimensional input space motivates the use of Deep Learning (DL) solutions with their capabilities of performing automatic feature extraction from raw data.

\paragraph*{Networks} The benchmark includes 3 convolutional neural networks (CNNs) of different complexity: different numbers of filters and different sizes of the input space. All networks contain only convolutional and fully connected layers with ReLU activations. All CNNs perform the regression function. They have been trained on the same dataset (time series data for mechanical component degradation during flight).

\paragraph*{Specifications} We propose 3 properties for the NN-based RUL estimation function. First, two properties (robustness and monotonicity) are local, i.e., defined around a given point. We provide a script with an adjustable random seed that can generate these properties around input points randomly picked from a test dataset. For robustness properties, the input perturbation (delta) is varied between 5\% and 40\%, while the number of perturbed inputs varies between 2 and 16. For monotonicity properties, monotonic shifts between 5\% and 20\% from a given point are considered. Properties of the last type ("if-then") require the output (RUL) to be in an expected value range given certain input ranges. Several if-then properties of different complexity are provided (depending on range widths).

\paragraph*{Link} \url{https://github.com/loonwerks/vnncomp2022}

\paragraph*{Paper} Available in~\cite{kirov2023benchmark} or on request.

\pagebreak
\subsection{VGGNET16 2023}
\paragraph*{Proposed by} Stanley Bak, Stony Brook University

\paragraph*{Motivation} This benchmark tries to scale up the size of networks being analyzed by using the well-studied VGGNET-16 architecture~\cite{simonyan2014very} that runs on ImageNet. Input-output properties are proposed on pixel-level perturbations that can lead to image misclassification. 

\paragraph*{Networks} All properties are run on the same network, which includes 138 million parameters. The network features convolution layers, ReLU activation functions, as well as max pooling layers.

\paragraph*{Specifications} Properties analyzed ranged from single-pixel perturbations to perturbations on all 150528 pixles (L-infinity perturbations). A subset of the images was used to create the specifications, one from each category, which was randomly chosen to attack. Pixels to perturb were also randomly selected according to a random seed.

\paragraph*{Link} \url{https://github.com/stanleybak/vggnet16_benchmark2022/}

\subsection{TLL Verify Bench}
\paragraph*{Proposed by} James Ferlez (University of California, Irvine)

\paragraph*{Motivation} This benchmark consists of Two-Level Lattice (TLL) NNs, which have been shown to be amenable to fast verification algorithms (e.g. \cite{FerlezKS22}). Thus, this benchmark was proposed as a means of comparing TLL-specific verification algorithms with general-purpose NN verification algorithms (i.e. algorithms that can verify arbitrary deep, fully-connected ReLU NNs).

\paragraph*{Networks}  The networks in this benchmark are a subset of the ones used in \cite[Experiment 3]{FerlezKS22}. Each of these TLL NNs has $n=2$ inputs and $m=1$ output. The architecture of a TLL NN is further specified by two parameters: $N$, the number of local linear functions, and $M$, the number of selector sets. This benchmark contains TLLs of sizes $N = M = 8, 16, 24, 32, 40, 48, 56, 64$, with $30$ randomly generated examples of each (the generation procedure is described in \cite[Section 6.1.1]{FerlezKS22}). At runtime, the specified verification timeout determines how many of these networks are included in the benchmark so as to achieve an overall 6-hour run time; this selection process is deterministic. Finally, a TLL NN has a natural representation using multiple computation paths \cite[Figure 1]{FerlezKS22}, but many tools are only compatible with fully-connected networks. Hence, the ONNX models in this benchmark implement TLL NNs by ``stacking'' these computation paths to make a fully connected NN (leading to sparse weight matrices: i.e. with many zero weights and biases). The \texttt{TLLnet} class (\url{https://github.com/jferlez/TLLnet}) contains the code necessary to generate these implementations via the \texttt{exportONNX} method.

\paragraph*{Specifications}  All specifications have as input constraints the hypercube $[-2,2]^2$. Since all networks have only a single output, the output properties consist of a randomly generated real number and a randomly generated inequality direction. Random output samples from the network are used to roughly ensure that the real number property has an equal likelihood of being within the output range of the NN and being outside of it (either above or below all NN outputs on the input constraint set). The inequality direction is generated independently and with each direction having an equal probability. This scheme biases the benchmark towards verification problems for which counterexamples exist. 

\paragraph*{Link} \url{https://github.com/jferlez/TLLVerifyBench}
\paragraph*{Commit}
199d2c26d0ec456e62906366b694a875a21ff7ef

\subsection{Traffic Signs Recognition}
\paragraph*{Proposed by} M\u{a}d\u{a}lina Era\c{s}cu and Andreea Postovan (West University of Timisoara, Romania)
\paragraph*{Motivation} Traffic signs play a crucial role in ensuring road safety and managing traffic flow in both city and highway driving. The recognition of these signs, a vital component of autonomous driving vision systems, faces challenges such as susceptibility to adversarial examples~\cite{szegedy2013intriguing} and occlusions~\cite{zhang2020lightweight}, stemming from diverse traffic scene conditions.

\paragraph*{Networks} Binary neural networks (BNNs) show promise in computationally limited and energy-constrained environments within the realm of autonomous driving~\cite{hubara2016binarized}. BNNs, where weights and/or activations are binarized to $\pm 1$, offer reduced model size and simplified convolution operations for image recognition compared to traditional neural networks (NNs).

We trained and tested various BNN architectures using the German Traffic Sign Recognition Benchmark (GTSRB) dataset~\cite{GTSRB}. This multi-class dataset, containing images of German road signs across 43 classes, poses challenges for both humans and models due to factors like perspective change, shade, color degradation, and lighting conditions. The dataset was also tested using the Belgian Traffic Signs \cite{BelgianTrafficSignDatabase} and Chinese Traffic Signs \cite{ChineseTrafficSignDatabase} datasets. The Belgium Traffic Signs dataset, with 62 classes, had 23 overlapping classes with GTSRB. The Chinese Traffic Signs dataset, with 58 classes, shared 15 classes with GTSRB. Pre-processing steps involved relabeling classes in the Belgium and Chinese datasets to match those in GTSRB and eliminating non-overlapping classes (see \cite{postovan2023architecturing} for details).

\begin{figure}[h]
  \centering
    \includegraphics[width=0.7\textwidth]{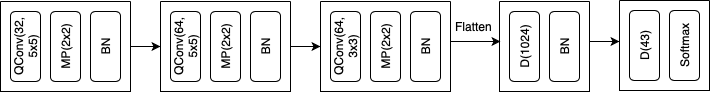}
    \caption{Accuracy Efficient Architecture for GTSRB and Belgium dataset}
    \label{fig:Acc-Efficient-Arch-GTSRB-Belgium}
\end{figure}

\begin{figure}[h]
  \centering
    \includegraphics[width=0.7\textwidth]{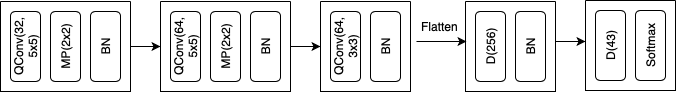}
    \caption{Accuracy Efficient Architecture for Chinese dataset}
    \label{fig:Acc-Efficient-Arch-Chinese}
\end{figure}

\begin{figure}[h]
  \centering
    \includegraphics[width=0.3\textwidth]{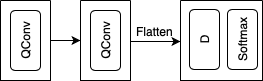}
    \caption{XNOR(QConv) architecture}
    \label{fig:XNOR(QConv)-arch}
\end{figure}

We provide three models with the structure in Figures \ref{fig:Acc-Efficient-Arch-GTSRB-Belgium}, \ref{fig:Acc-Efficient-Arch-Chinese}, and \ref{fig:XNOR(QConv)-arch}. They contain QConv, Batch Normalization (BN), Max Pooling (ML), Fully Connected/Dense (D) layers.  Note that the QConv layer binarizes the corresponding convolutional layer. All models were trained for 30 epochs. The model from Figure \ref{fig:Acc-Efficient-Arch-GTSRB-Belgium} was trained with images having the dimension 64px x 64 px, the one from Figure \ref{fig:Acc-Efficient-Arch-Chinese} with 48px x 48 px and the one from Figure \ref{fig:XNOR(QConv)-arch} with 30px x 30 px. The two models involving Batch Normalization layers introduce real valued parameters besides the binary ones, while the third one contains only binary parameters (see Table \ref{tab:stats}) for statistics.

\begin{table}[h]
\caption{Training and Testing Statistics}
\label{tab:stats}
\centering
\scriptsize
\begin{tabular}{|c|c|ccc|ccc|}
\hline
\multirow{2}{*}{\textbf{Input size}} & \multirow{2}{*}{\textbf{Model name}} & \multicolumn{3}{c|}{\textbf{Accuracy}}                                      & \multicolumn{3}{c|}{\textbf{\#Params}}                                      \\ \cline{3-8} 
                            &                             & \multicolumn{1}{c|}{\textbf{German}} & \multicolumn{1}{c|}{\textbf{China}} & \textbf{Belgium} & \multicolumn{1}{c|}{\textbf{Binary}}  & \multicolumn{1}{c|}{\textbf{Real}} & \textbf{Total}   \\ \hline
64px $\times$ 64px          & Figure \ref{fig:Acc-Efficient-Arch-GTSRB-Belgium}                  & \multicolumn{1}{c|}{96.45}  & \multicolumn{1}{c|}{81.50} & 88.17   & \multicolumn{1}{c|}{1772896} & \multicolumn{1}{c|}{2368} & 1775264 \\ \hline
48px $\times$ 48px          & Figure \ref{fig:Acc-Efficient-Arch-Chinese}                  & \multicolumn{1}{c|}{95.28}  & \multicolumn{1}{c|}{83.90} & 87.78   & \multicolumn{1}{c|}{904288}  & \multicolumn{1}{c|}{832}  & 905120  \\ \hline
30px $\times$ 30px          & Figure \ref{fig:XNOR(QConv)-arch}                  & \multicolumn{1}{c|}{81.54}  & \multicolumn{1}{c|}{N/A}   & N/A     & \multicolumn{1}{c|}{1005584} & \multicolumn{1}{c|}{0}    & 1005584 \\ \hline
\end{tabular}
\end{table}
\paragraph*{Specifications} To evaluate the \emph{adversarial robustness} of the networks above, we assessed perturbations within the infinity norm around zero, with the radius denoted as $\epsilon = \{1, 3, 5, 10, 15\}$. This involved randomly selecting three distinct images from the GTSRB dataset's test set for each model and generating \textsc{VNNLIB} files for each epsilon in the set. In total, we created 45 \textsc{VNNLIB} files. Due to a 6-hour total timeout constraint for solving all instances, each instance had a maximum timeout of 480 seconds. To review the generated \textsc{VNNLIB} specification files submitted to VNNCOMP 2023, as well as to generate new ones, please refer to \url{https://github.com/apostovan21/vnncomp2023}.

\paragraph*{Link} \url{https://github.com/apostovan21/vnncomp2023}

\subsection{ViT}
\paragraph*{Proposed by} the $\alpha,\!\beta$-CROWN team.
\paragraph*{Motivation}
Transformers~\cite{vaswani2017attention} based on the self-attention mechanism have much more complicated architectures and contain more kinds of nonlinerities, compared to simple feedforward networks with relatively simple activation functions. 
It makes verifying Transformers challenging. We aim to encourage the development of verification techniques for Transformer-based models, and we also aim to benchmark neural network verifiers on relatively complicated neural network architectures and more general nonlinearities. Therefore, we propose a new benchmark with Vision Transformers (ViTs)~\cite{dosovitskiy2020image}. This benchmark is developed based on our work on neural network verification for models with general nonlinearities~\cite{shi2023formal}.

\paragraph*{Networks}
The benchmark contains two ViTs, as shown in \Cref{tab:vits}.
Considering the difficulty of verifying ViTs, we modify the ViTs and make the models relatively shallow and narrow, with significantly reduced number of layers and attention heads.
Following \cite{shi2019robustness}, we also replace the layer normalization with batch normalization.
The models are mainly trained with PGD training~\cite{madry2017towards}, and we also add a weighted IBP~\cite{gowal2018effectiveness,shi2021fast} loss for one of the models as a regularization.

\begin{table}[ht]
\centering
\caption{Networks in the ViT benchmark.}
\label{tab:vits}
\begin{tabular}{ccc}
\toprule 
Model & \texttt{PGD\_2\_3\_16} & \texttt{IBP\_3\_3\_8} \\
\midrule
Layers & 2 & 3\\
Attention heads & 3 & 3\\
Patch size & 16 & 8\\
Weight of IBP loss & 0 & 0.01\\
Training $\epsilon$ & $\frac{2}{255}$ & $\frac{1}{255}$\\
Clean accuracy & 59.78\% & 62.21\%\\
\bottomrule
\end{tabular}
\end{table}

\paragraph*{Specifications} 
The specifications are generated from the robustness verification problem with $\ell_\infty$ perturbation. 
We use the CIFAR-10 dataset with perturbation size $\epsilon=\frac{1}{255}$ at test time.
We have filtered the CIFAR-10 test set to exclude instances where either adversarial examples can be found or the vanilla CROWN-like method~\cite{zhang2018efficient,shi2019robustness} can already easily verify.
We randomly keep 100 instances for each model, with a timeout threshold of 100 seconds.

\paragraph*{Link} \url{https://github.com/shizhouxing/ViT_vnncomp2023}

\subsection{Real-world distribution shifts}
\paragraph*{Proposed by} the Marabou team.
\paragraph*{Motivation}
While robustness against handcrafted perturbations (e.g., norm-bounded) for perception networks are more commonly investigated, robustness against real-world distribution shifts~\cite{wu2022toward} are less studied but of practical interests. This benchmark set contains queries for verifying the latter type of robustness.  
\paragraph*{Networks} The network is a concatenation of a generative model and a MNIST classifier. The generative model is trained to take in an unperturbed image and an embedding of a particular type of distribution shifts in latent space, and produce a perturbed image. The distribution shift captured in this case is the "shear" perturbation. 
\paragraph*{Specifications} The verification task is to certify that a classifier correctly classifies all images in a perturbation set, which is a set of images generated by the generative model given a fixed image and a ball centering the mean perturbations on this image (in the latent space). This mean perturbation is computed by a prior network.
\paragraph*{Link} \url{https://github.com/wu-haoze/dist-shift-vnn-comp}

\subsection{ml4acopf}
\paragraph*{Proposed by} Haoruo Zhao, Michael Klaminkin, Mathieu Tanneau, Wenbo Chen, and Pascal Van Hentenryck (Georgia Institute of Technology), and Hassan Hijazi, Juston Moore, and Haydn Jones (Los Alamos National Laboratory).

\paragraph*{Motivation}
Machine learning models are utilized to predict solutions for an optimization model known as AC Optimal Power Flow (ACOPF) in the power system. Since the solutions are continuous, a regression model is employed. The objective is to evaluate the quality of these machine learning model predictions, specifically by determining whether they satisfy the constraints of the optimization model. Given the challenges in meeting some constraints, the goal is to verify whether the worst-case violations of these constraints are within an acceptable tolerance level.

\paragraph*{Networks}
The neural network designed comprises two components. The first component predicts the solutions of the optimization model, while the second evaluates the violation of each constraint that needs checking. The first component consists solely of general matrix multiplication (GEMM) and rectified linear unit (ReLU) operators. However, the second component has a more complex structure, as it involves evaluating the violation of AC constraints using nonlinear functions, including sigmoid, quadratic, and trigonometric functions such as sine and cosine. This complex evaluation component is incorporated into the network due to a limitation of the VNNLIB format, which does not support trigonometric functions. Therefore, these constraints violation evaluation are included in the neural network.

\paragraph*{Specifications}
In this benchmark, four different properties are checked, each corresponding to a type of constraint violation:
\begin{enumerate}
    \item Power balance constraints: the net power at each bus node is equal to the sum of the power flows in the branches connected to that node.
    \item Thermal limit constraints: power flow on a transmission line is within its maximum and minimum limits.
    \item Generation bounds: a generator's active and reactive power output is within its maximum and minimum limits.
    \item Voltage magnitude bounds: a voltage's magnitude output is within its maximum and minimum limits.
\end{enumerate}

The input to the model is the active and reactive load. The chosen input point for perturbation is a load profile for which a corresponding feasible solution to the ACOPF problem is known to exist. For the feasibility check, the input load undergoes perturbation. Although this perturbation does not exactly match physical laws, the objective is to ascertain whether a machine learning-predicted solution with the perturbation can produce a solution that does not significantly violate the constraints.

The scale of the perturbation and the violation threshold are altered by testing whether an adversarial example can be easily found using projected gradient descent with the given perturbation. The benchmark, provided with a fixed random seed, is robust against the simple projected gradient descent that is implemented.

\paragraph*{Link} \url{https://github.com/AI4OPT/ml4acopf_benchmark}

\subsection{ACAS Xu}
\paragraph{Networks} The ACASXu benchmark consists of ten properties defined over 45 neural networks used to issue turn advisories to aircraft to avoid collisions. The neural networks have 300 neurons arranged in 6 layers, with ReLU activation functions. There are five inputs corresponding to the aircraft states, and five network outputs, where the minimum output is used as the turn advisory the system ultimately produces.

\paragraph{Specifications} We use the original 10 properties~\cite{katz2017reluplex}, where properties 1-4 are checked on all 45 networks as was done in later work by the original authors~\cite{katz2019marabou}. Properties 5-10 are checked on a single network. The total number of benchmarks is therefore 186. The original verification times ranged from seconds to days---including some benchmark instances that did not finish. This year we used a timeout of around two minutes (116 seconds) for each property, in order to fit within a total maximum runtime of six hours.





\newpage
\section{Results}
\label{sec:results}

Each tool was run on each of the benchmarks and produced a \texttt{csv} result file, that was provided as feedback to the tool authors using the online execution platform.
The final \texttt{csv} files for each tool as well as scoring scripts are available online: \url{https://github.com/ChristopherBrix/vnncomp2023_results}.
The results were analyzed automatically to compute scores and create the statistics presented in this section.

%
%
Penalties usually occur when a tool produces an incorrect result.

\begin{table}[h]
\begin{center}
\caption{Overall Score} \label{tab:score}
{\setlength{\tabcolsep}{2pt}
\begin{tabular}[h]{@{}lll@{}}
\toprule
\textbf{\# ~} & \textbf{Tool} & \textbf{Score}\\
\midrule
1 & $\alpha$-$\beta$-CROWN & 930.9 \\
2 & Marabou & 594.1 \\
3 & PyRAT & 585.5 \\
4 & NeuralSAT & 547.0 \\
5 & nnenum & 441.9 \\
6 & NNV & 176.4 \\
7 & FastBATLLNN & 100.0 \\
\bottomrule
\end{tabular}
}
\end{center}
\end{table}

\begin{figure}[h]
\centerline{\includegraphics[width=\textwidth]{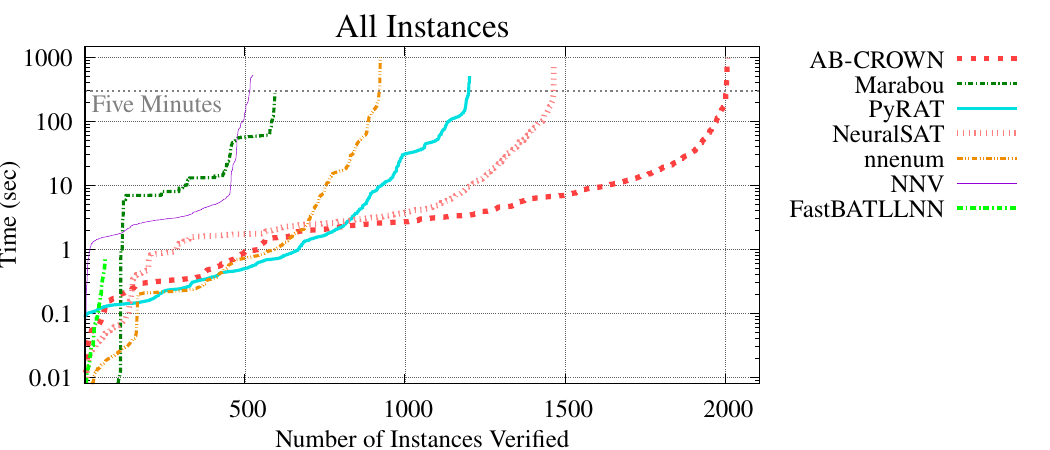}}
\caption{Cactus Plot for All Instances.}
\label{fig:quantPic}
\end{figure}

\begin{figure}[h]
\centerline{\includegraphics[width=\textwidth]{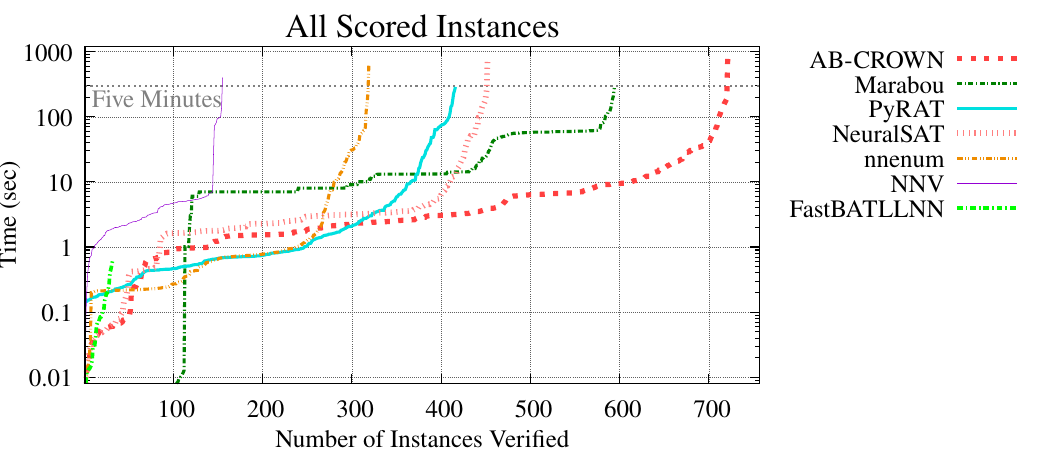}}
\caption{Cactus Plot for All Scored Instances.}
\label{fig:quantPic}
\end{figure}

\clearpage
\subsection{Other Stats}

This section presents other statistics related to the measurements that are interesting but did not play a direct role in scoring this year.



\begin{table}[h]
\begin{center}
\caption{Overhead} \label{tab:overhead}
{\setlength{\tabcolsep}{2pt}
\begin{tabular}[h]{@{}llr@{}}
\toprule
\textbf{\# ~} & \textbf{Tool} & \textbf{Seconds}\\
\midrule
1 & FastBATLLNN & 0.3 \\
2 & nnenum & 0.9 \\
3 & Marabou & 1.1 \\
4 & PyRAT & 3.4 \\
5 & NeuralSAT & 3.6 \\
6 & $\alpha$-$\beta$-CROWN & 5.7 \\
7 & NNV & 15.9 \\
\bottomrule
\end{tabular}
}
\end{center}
\end{table}


\begin{table}[h]
\begin{center}
\caption{Num Benchmarks Participated} \label{tab:stats0}
{\setlength{\tabcolsep}{2pt}
\begin{tabular}[h]{@{}llr@{}}
\toprule
\textbf{\# ~} & \textbf{Tool} & \textbf{Count}\\
\midrule
1 & $\alpha$-$\beta$-CROWN & 10 \\
2 & Marabou & 9 \\
3 & PyRAT & 8 \\
4 & NeuralSAT & 8 \\
5 & nnenum & 6 \\
6 & NNV & 6 \\
7 & FastBATLLNN & 1 \\
\bottomrule
\end{tabular}
}
\end{center}
\end{table}


\begin{table}[h]
\begin{center}
\caption{Num Instances Verified} \label{tab:stats1}
{\setlength{\tabcolsep}{2pt}
\begin{tabular}[h]{@{}llr@{}}
\toprule
\textbf{\# ~} & \textbf{Tool} & \textbf{Count}\\
\midrule
1 & $\alpha$-$\beta$-CROWN & 721 \\
2 & Marabou & 594 \\
3 & NeuralSAT & 452 \\
4 & PyRAT & 416 \\
5 & nnenum & 319 \\
6 & NNV & 155 \\
7 & FastBATLLNN & 32 \\
\bottomrule
\end{tabular}
}
\end{center}
\end{table}


\begin{table}[h]
\begin{center}
\caption{Num SAT} \label{tab:stats2}
{\setlength{\tabcolsep}{2pt}
\begin{tabular}[h]{@{}llr@{}}
\toprule
\textbf{\# ~} & \textbf{Tool} & \textbf{Count}\\
\midrule
1 & $\alpha$-$\beta$-CROWN & 151 \\
2 & Marabou & 128 \\
3 & PyRAT & 109 \\
4 & NeuralSAT & 107 \\
5 & nnenum & 101 \\
6 & NNV & 60 \\
7 & FastBATLLNN & 17 \\
\bottomrule
\end{tabular}
}
\end{center}
\end{table}


\begin{table}[h]
\begin{center}
\caption{Num UNSAT} \label{tab:stats3}
{\setlength{\tabcolsep}{2pt}
\begin{tabular}[h]{@{}llr@{}}
\toprule
\textbf{\# ~} & \textbf{Tool} & \textbf{Count}\\
\midrule
1 & $\alpha$-$\beta$-CROWN & 570 \\
2 & Marabou & 466 \\
3 & NeuralSAT & 345 \\
4 & PyRAT & 307 \\
5 & nnenum & 218 \\
6 & NNV & 95 \\
7 & FastBATLLNN & 15 \\
\bottomrule
\end{tabular}
}
\end{center}
\end{table}


\begin{table}[h]
\begin{minipage}{\textwidth}
\begin{center}
\caption{Incorrect Results (or Missing CE)} \label{tab:stats4}
{\setlength{\tabcolsep}{2pt}
\begin{tabular}[h]{@{}llr@{}}
\toprule
\textbf{\# ~} & \textbf{Tool} & \textbf{Count}\\
\midrule
1 & NeuralSAT\footnote{All penalties were due to incorrectly parsed constraints in the nn4sys benchmark.} & 35 \\
2 & NNV & 35 \\
3 & $\alpha$-$\beta$-CROWN & 1 \\
\bottomrule
\end{tabular}
}
\end{center}
\end{minipage}
\end{table}

\section{Conclusion and Ideas for Future Competitions}
\label{sec:conclusion}
This report summarizes the 4$^\text{rd}$ Verification of Neural Networks Competition (VNN-COMP), held in 2023.
While we observed a significant increase in the diversity, complexity, and scale of the proposed benchmarks, the best-performing tools seem to converge to GPU-enabled linear bound propagation methods using a branch-and-bound framework.
In addition to the standardization of input formats (\texttt{onnx} and \texttt{vnnlib}) and evaluation hardware, introduced for VNN-COMP 2021, VNN-COMP 2023 also continued the standardized format for counter-examples and fully automated evaluation pipeline introduced in VNN-COMP 2022, requiring authors to provide complete installation scripts.
We hope that this increased standardization and automatization does not only simplify the evaluation during the competition but also enables practitioners and researchers to more easily apply a range of state-of-the-art verification methods to their individual problems.

VNN-COMP 2023, successfully implemented a range of improvement opportunities identified during the previous iteration. These included requiring witnesses of found counter-examples to disambiguate tool disagreement, increasing automatization to enable a smoother final evaluation, and making a broader range of AWS instances available to allow for a better fit with tools' requirements. 
Further ideas for future competitions include the use of scored benchmarks specifically designed for year-on-year progress tracking, the reduction of tool tuning, a batch-processing mode, and more rigorous soundness evaluation.

\section*{Acknowledgements}
%


This research was supported in part by the Air Force Research Laboratory Information Directorate, through the Air Force Office of Scientific Research Summer Faculty Fellowship
Program, Contract Numbers FA8750-15-3-6003, FA9550-15-0001 and FA9550-20-F-0005.
This material is based upon work supported by the Air Force Office of Scientific Research under award numbers FA9550-19-1-0288, FA9550-21-1-0121, and FA9550-22-1-0019, the National Science Foundation (NSF) under grant numbers 1918450, 1911017, 2028001, 2107035, 2220401, 2220418, 2220426, and 2238133, and the Defense Advanced Research Projects Agency (DARPA) Assured Autonomy and Assured Neuro Symbolic Learning and Reasoning (ANSR)
programs through contract numbers FA8750-18-C-0089 and FA8750-23-C-0518.
Any opinions, findings, and conclusions or recommendations expressed in this material are those of the author(s) and do not necessarily reflect the views of the United States Air Force, DARPA, nor NSF.

Tool and benchmark authors listed in \Cref{sec:participants} and \Cref{sec:benchmarks} participated in the preparation and review of this report.

\clearpage
\label{sect:bib}
\bibliographystyle{plain}
\bibliography{bib/nnv, bib/nnenum, bib/peregriNN, bib/verinet, bib/oval,bib/venus,bib/MIPVerify, bib/mnbab, bib/alpha-beta-CROWN, bib/collins,bib/dnnf,bib/nvjl,bib/nn4sys,bib/Marabou,bib/RPM,bib/AVeriNN,bib/VeRAPAk,bib/general,bib/traffic-signs-recognition,bib/neuralsat, bib/pyrat, bib/vit}

\begin{thebibliography}{10}

\bibitem{BelgianTrafficSignDatabase}
{Belgian Traffic Sign Database}.
\newblock \url{https://www.kaggle.com/datasets/shazaelmorsh/trafficsigns}.
\newblock Accessed: March 25th, 2023.

\bibitem{ChineseTrafficSignDatabase}
{Chinese Traffic Sign Database}.
\newblock
  \url{https://www.kaggle.com/datasets/dmitryyemelyanov/chinese-traffic-signs}.
\newblock Accessed: March 25th, 2023.

\bibitem{GTSRB}
{German Traffic Sign Recognition Benchmark}.
\newblock
  \url{https://www.kaggle.com/datasets/meowmeowmeowmeowmeow/gtsrb-german-traffic-sign?datasetId=82373&language=Python}.
\newblock Accessed: March 25th, 2023.

\bibitem{pyrat-website}
{PyRAT Analyzer website}.
\newblock \url{https://pyrat-analyzer.com/}.
\newblock Accessed: December 15th, 2023.

\bibitem{bak2020vnn}
Stanley Bak.
\newblock Execution-guided overapproximation (ego) for improving scalability of
  neural network verification, 2020.

\bibitem{bak2021nnenum}
Stanley Bak.
\newblock nnenum: Verification of relu neural networks with optimized
  abstraction refinement.
\newblock In {\em NASA Formal Methods Symposium}, pages 19--36. Springer, 2021.

\bibitem{bak2020cav}
Stanley Bak, Hoang-Dung Tran, Kerianne Hobbs, and Taylor~T. Johnson.
\newblock Improved geometric path enumeration for verifying {ReLU} neural
  networks.
\newblock In {\em 32nd International Conference on Computer-Aided Verification
  (CAV)}, July 2020.

\bibitem{bunelunified2018}
Rudy Bunel, Ilker Turkaslan, Philip~HS Torr, Pushmeet Kohli, and M~Pawan Kumar.
\newblock A unified view of piecewise linear neural network verification.
\newblock {\em Advances in Neural Information Processing Systems}, 2018.

\bibitem{vnnlib}
Stefano Demarchi, Dario Guidotti, Luca Pulina, and Armando Tacchella.
\newblock Supporting standardization of neural networks verification with
  vnnlib and coconet.
\newblock In Nina Narodytska, Guy Amir, Guy Katz, and Omri Isac, editors, {\em
  Proceedings of the 6th Workshop on Formal Methods for ML-Enabled Autonomous
  Systems}, volume~16 of {\em Kalpa Publications in Computing}, pages 47--58.
  EasyChair, 2023.

\bibitem{dosovitskiy2020image}
Alexey Dosovitskiy, Lucas Beyer, Alexander Kolesnikov, Dirk Weissenborn,
  Xiaohua Zhai, Thomas Unterthiner, Mostafa Dehghani, Matthias Minderer, Georg
  Heigold, Sylvain Gelly, et~al.
\newblock An image is worth 16x16 words: Transformers for image recognition at
  scale.
\newblock In {\em International Conference on Learning Representations}, 2020.

\bibitem{duong2023dpllt}
Hai Duong, Linhan Li, ThanhVu Nguyen, and Matthew Dwyer.
\newblock {A DPLL(T) Framework for Verifying Deep Neural Networks}, 2023.
\newblock arXiv, 25 pages.

\bibitem{durand2022reciph}
Serge Durand, Augustin Lemesle, Zakaria Chihani, Caterina Urban, and
  Fran{\c{c}}ois Terrier.
\newblock Reciph: Relational coefficients for input partitioning heuristic.
\newblock In {\em 1st Workshop on Formal Verification of Machine Learning
  (WFVML 2022)}, 2022.

\bibitem{ForMuLA}
{EASA and Collins Aerospace}.
\newblock {Formal Methods use for Learning Assurance (ForMuLA)}.
\newblock Technical report, April 2023.

\bibitem{FerlezKS22}
James Ferlez, Haitham Khedr, and Yasser Shoukry.
\newblock Fast {BATLLNN:} fast box analysis of two-level lattice neural
  networks.
\newblock In Ezio Bartocci and Sylvie Putot, editors, {\em {HSCC} '22: 25th
  {ACM} International Conference on Hybrid Systems: Computation and Control,
  Milan, Italy, May 4 - 6, 2022}, pages 23:1--23:11. {ACM}, 2022.

\bibitem{gowal2018effectiveness}
Sven Gowal, Krishnamurthy Dvijotham, Robert Stanforth, Rudy Bunel, Chongli Qin,
  Jonathan Uesato, Relja Arandjelovic, Timothy Mann, and Pushmeet Kohli.
\newblock On the effectiveness of interval bound propagation for training
  verifiably robust models.
\newblock {\em arXiv preprint arXiv:1810.12715}, 2018.

\bibitem{he2022Characterizing}
Haoyu He, Tianhao Wei, Huan Zhang, Changliu Liu, and Cheng Tan.
\newblock Characterizing neural network verification for systems with
  {NN4SYSBench}.
\newblock {\em 1st Workshop on Formal Verification of Machine Learning}, 2022.

\bibitem{hubara2016binarized}
Itay Hubara, Matthieu Courbariaux, Daniel Soudry, Ran El-Yaniv, and Yoshua
  Bengio.
\newblock {Binarized Neural Networks}.
\newblock {\em Advances in Neural Information Processing Systems}, 29, 2016.

\bibitem{katz2017reluplex}
Guy Katz, Clark Barrett, David~L Dill, Kyle Julian, and Mykel~J Kochenderfer.
\newblock Reluplex: An efficient smt solver for verifying deep neural networks.
\newblock In {\em International Conference on Computer Aided Verification},
  pages 97--117. Springer, 2017.

\bibitem{KatzHIJLLSTWZDK19}
Guy Katz, Derek~A. Huang, Duligur Ibeling, Kyle Julian, Christopher Lazarus,
  Rachel Lim, Parth Shah, Shantanu Thakoor, Haoze Wu, Aleksandar Zeljic,
  David~L. Dill, Mykel~J. Kochenderfer, and Clark~W. Barrett.
\newblock The marabou framework for verification and analysis of deep neural
  networks.
\newblock In Isil Dillig and Serdar Tasiran, editors, {\em Computer Aided
  Verification - 31st International Conference, {CAV} 2019, New York City, NY,
  USA, July 15-18, 2019, Proceedings, Part {I}}, volume 11561 of {\em Lecture
  Notes in Computer Science}, pages 443--452. Springer, 2019.

\bibitem{katz2019marabou}
Guy Katz, Derek~A Huang, Duligur Ibeling, Kyle Julian, Christopher Lazarus,
  Rachel Lim, Parth Shah, Shantanu Thakoor, Haoze Wu, Aleksandar Zelji{\'c},
  et~al.
\newblock The marabou framework for verification and analysis of deep neural
  networks.
\newblock In {\em International Conference on Computer Aided Verification},
  pages 443--452. Springer, 2019.

\bibitem{kirov2023benchmark}
Dmitrii Kirov and Simone~Fulvio Rollini.
\newblock Benchmark: remaining useful life predictor for aircraft equipment.
\newblock In {\em International Conference on Bridging the Gap between AI and
  Reality}, pages 299--304. Springer, 2023.

\bibitem{kirov2023formal}
Dmitrii Kirov, Simone~Fulvio Rollini, Luigi Di~Guglielmo, and Darren Cofer.
\newblock Formal verification of a neural network based prognostics system for
  aircraft equipment.
\newblock In {\em International Conference on Bridging the Gap between AI and
  Reality}, pages 225--240. Springer, 2023.

\bibitem{kraska18case}
Tim Kraska, Alex Beutel, Ed~H Chi, Jeffrey Dean, and Neoklis Polyzotis.
\newblock The case for learned index structures.
\newblock In {\em Proceedings of the 2018 International Conference on
  Management of Data}, 2018.

\bibitem{manzanas2023cav}
Diego~Manzanas Lopez, Sung~Woo Choi, Hoang-Dung Tran, and Taylor~T. Johnson.
\newblock {NNV 2.0}: The neural network verification tool.
\newblock In {\em 35th International Conference on Computer-Aided Verification
  (CAV)}, July 2023.

\bibitem{madry2017towards}
Aleksander Madry, Aleksandar Makelov, Ludwig Schmidt, Dimitris Tsipras, and
  Adrian Vladu.
\newblock Towards deep learning models resistant to adversarial attacks.
\newblock {\em arXiv preprint arXiv:1706.06083}, 2017.

\bibitem{manzanas2022formats}
Diego Manzanas~Lopez, Patrick Musau, Nathaniel Hamilton, and Taylor Johnson.
\newblock Reachability analysis of a general class of neural ordinary
  differential equation.
\newblock In {\em Proceedings of the 20th International Conference on Formal
  Modeling and Analysis of Timed Systems (FORMATS 2022), Co-Located with
  CONCUR, FMICS, and QEST as part of CONFEST 2022.}, Warsaw, Poland, September
  2022.

\bibitem{postovan2023architecturing}
Andreea Postovan and M{\u{a}}d{\u{a}}lina Era{\c{s}}cu.
\newblock Architecturing binarized neural networks for traffic sign
  recognition.
\newblock {\em arXiv preprint arXiv:2303.15005}, 2023.

\bibitem{shi2023formal}
Zhouxing Shi, Qirui Jin, Huan Zhang, Zico Kolter, Suman Jana, and Cho-Jui
  Hsieh.
\newblock Formal verification for neural networks with general nonlinearities
  via branch-and-bound.
\newblock In {\em 2nd Workshop on Formal Verification of Machine Learning
  (WFVML 2023)}, 2023.

\bibitem{shi2021fast}
Zhouxing Shi, Yihan Wang, Huan Zhang, Jinfeng Yi, and Cho-Jui Hsieh.
\newblock Fast certified robust training with short warmup.
\newblock {\em Advances in Neural Information Processing Systems},
  34:18335--18349, 2021.

\bibitem{shi2019robustness}
Zhouxing Shi, Huan Zhang, Kai-Wei Chang, Minlie Huang, and Cho-Jui Hsieh.
\newblock Robustness verification for transformers.
\newblock In {\em International Conference on Learning Representations}, 2019.

\bibitem{simonyan2014very}
Karen Simonyan and Andrew Zisserman.
\newblock Very deep convolutional networks for large-scale image recognition.
\newblock {\em arXiv preprint arXiv:1409.1556}, 2014.

\bibitem{DeepPoly:19}
Gagandeep Singh, Timon Gehr, Markus P{\"{u}}schel, and Martin~T. Vechev.
\newblock An abstract domain for certifying neural networks.
\newblock {\em Proc. {ACM} Program. Lang.}, 3({POPL}):41:1--41:30, 2019.

\bibitem{szegedy2013intriguing}
Christian Szegedy, Wojciech Zaremba, Ilya Sutskever, Joan Bruna, Dumitru Erhan,
  Ian Goodfellow, and Rob Fergus.
\newblock {Intriguing Properties of Neural Networks}.
\newblock {\em arXiv preprint arXiv:1312.6199}, 2013.

\bibitem{Tjeng2019EvaluatingRO}
Vincent Tjeng, Kai~Y. Xiao, and Russ Tedrake.
\newblock Evaluating robustness of neural networks with mixed integer
  programming.
\newblock In {\em ICLR}, 2019.

\bibitem{tran2021fac}
H.~D. Tran, N.~Pal, D.~Lopez, P.~Musau, X.~Yang, W.~Xiang L.~Nguyen, S.~Bak, ,
  and T.~T. Johnson.
\newblock Verification of piecewise deep neural networks: A star set approach
  with zonotope pre-filter.
\newblock {\em Formal aspects of computing}, 2021.

\bibitem{tran2020cav}
Hoang-Dung Tran, Stanley Bak, Weiming Xiang, and Taylor~T. Johnson.
\newblock Verification of deep convolutional neural networks using imagestars.
\newblock In {\em 32nd International Conference on Computer-Aided Verification
  (CAV)}. Springer, July 2020.

\bibitem{tran2019emsoft}
Hoang-Dung Tran, Feiyang Cei, Diego~Manzanas Lopez, Taylor~T. Johnson, and
  Xenofon Koutsoukos.
\newblock Safety verification of cyber-physical systems with reinforcement
  learning control.
\newblock In {\em ACM SIGBED International Conference on Embedded Software
  (EMSOFT'19)}. ACM, October 2019.

\bibitem{tran2023hscc}
Hoang~Dung Tran, SungWoo Choi, Tomoya Yamaguchi, Bardh Hoxha, and Danil
  Prokhorov.
\newblock Verification of recurrent neural networks using star reachability.
\newblock In {\em The 26th ACM International Conference on Hybrid Systems:
  Computation and Control (HSCC)}, May 2023.

\bibitem{tran2019fm}
Hoang-Dung Tran, Patrick Musau, Diego~Manzanas Lopez, Xiaodong Yang, Luan~Viet
  Nguyen, Weiming Xiang, and Taylor~T. Johnson.
\newblock Star-based reachability analysis for deep neural networks.
\newblock In {\em 23rd International Symposium on Formal Methods (FM'19)}.
  Springer International Publishing, October 2019.

\bibitem{tran2021cav}
Hoang-Dung Tran, Neelanjana Pal, Patrick Musau, Xiaodong Yang, Nathaniel~P.
  Hamilton, Diego~Manzanas Lopez, Stanley Bak, and Taylor~T. Johnson.
\newblock Robustness verification of semantic segmentation neural networks
  using relaxed reachability.
\newblock In {\em 33rd International Conference on Computer-Aided Verification
  (CAV)}. Springer, July 2021.

\bibitem{tran2020cav_tool}
Hoang-Dung Tran, Xiaodong Yang, Diego~Manzanas Lopez, Patrick Musau, Luan~Viet
  Nguyen, Weiming Xiang, Stanley Bak, and Taylor~T. Johnson.
\newblock {NNV}: The neural network verification tool for deep neural networks
  and learning-enabled cyber-physical systems.
\newblock In {\em 32nd International Conference on Computer-Aided Verification
  (CAV)}, July 2020.

\bibitem{vaswani2017attention}
Ashish Vaswani, Noam Shazeer, Niki Parmar, Jakob Uszkoreit, Llion Jones,
  Aidan~N Gomez, {\L}ukasz Kaiser, and Illia Polosukhin.
\newblock Attention is all you need.
\newblock {\em Advances in neural information processing systems}, 30, 2017.

\bibitem{wang2021betacrown}
Shiqi Wang, Huan Zhang, Kaidi Xu, Xue Lin, Suman Jana, Cho-Jui Hsieh, and Zico
  Kolter.
\newblock {Beta-CROWN}: Efficient bound propagation with per-neuron split
  constraints for complete and incomplete neural network verification.
\newblock {\em arXiv preprint arXiv:2103.06624}, 2021.

\bibitem{wei2023convex}
Dennis Wei, Haoze Wu, Min Wu, Pin-Yu Chen, Clark Barrett, and Eitan Farchi.
\newblock Convex bounds on the softmax function with applications to robustness
  verification.
\newblock In {\em International Conference on Artificial Intelligence and
  Statistics}, pages 6853--6878. PMLR, 2023.

\bibitem{vegas}
Haoze Wu, Clark Barrett, Mahmood Sharif, Nina Narodytska, and Gagandeep Singh.
\newblock Scalable verification of gnn-based job schedulers.
\newblock 6(OOPSLA2), oct 2022.

\bibitem{wu2020parallelization}
Haoze Wu, Alex Ozdemir, Aleksandar Zeljic, Kyle Julian, Ahmed Irfan, Divya
  Gopinath, Sadjad Fouladi, Guy Katz, Corina Pasareanu, and Clark Barrett.
\newblock Parallelization techniques for verifying neural networks.
\newblock In {\em \# PLACEHOLDER\_PARENT\_METADATA\_VALUE\#}, volume~1, pages
  128--137. TU Wien Academic Press, 2020.

\bibitem{wu2022toward}
Haoze Wu, Teruhiro Tagomori, Alexander Robey, Fengjun Yang, Nikolai Matni,
  George Pappas, Hamed Hassani, Corina Pasareanu, and Clark Barrett.
\newblock Toward certified robustness against real-world distribution shifts.
\newblock {\em arXiv preprint arXiv:2206.03669}, 2022.

\bibitem{wu2022efficient}
Haoze Wu, Aleksandar Zelji{\'c}, Guy Katz, and Clark Barrett.
\newblock Efficient neural network analysis with sum-of-infeasibilities.
\newblock In {\em International Conference on Tools and Algorithms for the
  Construction and Analysis of Systems}, pages 143--163. Springer, 2022.

\bibitem{xu2020automatic}
Kaidi Xu, Zhouxing Shi, Huan Zhang, Yihan Wang, Kai-Wei Chang, Minlie Huang,
  Bhavya Kailkhura, Xue Lin, and Cho-Jui Hsieh.
\newblock Automatic perturbation analysis for scalable certified robustness and
  beyond.
\newblock {\em Advances in Neural Information Processing Systems}, 33, 2020.

\bibitem{xu2021fast}
Kaidi Xu, Huan Zhang, Shiqi Wang, Yihan Wang, Suman Jana, Xue Lin, and Cho-Jui
  Hsieh.
\newblock {Fast and Complete}: Enabling complete neural network verification
  with rapid and massively parallel incomplete verifiers.
\newblock In {\em International Conference on Learning Representations}, 2021.

\bibitem{zhang2022general}
Huan Zhang*, Shiqi Wang*, Kaidi Xu*, Linyi Li, Bo~Li, Suman Jana, Cho-Jui
  Hsieh, and J~Zico Kolter.
\newblock General cutting planes for bound-propagation-based neural network
  verification.
\newblock {\em Advances in Neural Information Processing Systems (NeurIPS)},
  2022.

\bibitem{zhang2018efficient}
Huan Zhang, Tsui-Wei Weng, Pin-Yu Chen, Cho-Jui Hsieh, and Luca Daniel.
\newblock Efficient neural network robustness certification with general
  activation functions.
\newblock {\em Advances in Neural Information Processing Systems},
  31:4939--4948, 2018.

\bibitem{zhang2020lightweight}
Jianming Zhang, Wei Wang, Chaoquan Lu, Jin Wang, and Arun~Kumar Sangaiah.
\newblock {Lightweight Deep Network for Traffic Sign Classification}.
\newblock {\em Annals of Telecommunications}, 75:369--379, 2020.

\end{thebibliography}


\appendix

\clearpage
\section{Extended Results}
In this section, we provide more fine-grained results.

\subsection{Scored Benchmarks}
\label{sec:benchmark_results_scored}


\begin{table}[h]
\begin{center}
\caption{Benchmark \texttt{2023-acasxu}} \label{tab:cat_{cat}}
{\setlength{\tabcolsep}{2pt}
\begin{tabular}[h]{@{}llllllrr@{}}
\toprule
\textbf{\# ~} & \textbf{Tool} & \textbf{Verified} & \textbf{Falsified} & \textbf{Fastest} & \textbf{Penalty} & \textbf{Score} & \textbf{Percent}\\
\midrule
1 & nnenum & 139 & 47 & 0 & 0 & 1860 & 100.0\% \\
2 & NeuralSAT & 138 & 46 & 0 & 0 & 1840 & 98.9\% \\
3 & PyRAT & 137 & 45 & 0 & 0 & 1820 & 97.8\% \\
4 & Marabou & 135 & 46 & 0 & 0 & 1810 & 97.3\% \\
5 & $\alpha$-$\beta$-CROWN & 139 & 46 & 0 & 1 & 1700 & 91.4\% \\
6 & NNV & 68 & 27 & 0 & 0 & 950 & 51.1\% \\
\bottomrule
\end{tabular}
}
\end{center}
\end{table}

\begin{figure}[h]
\centerline{\includegraphics[width=\textwidth]{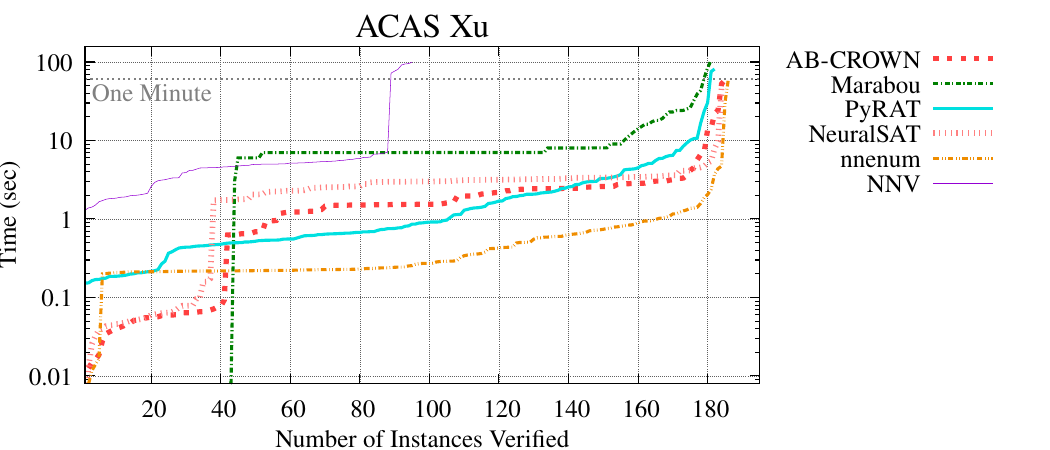}}
\caption{Cactus Plot for ACAS Xu.}
\label{fig:quantPic}
\end{figure}


\begin{table}[h]
\begin{center}
\caption{Benchmark \texttt{2023-cgan}} \label{tab:cat_{cat}}
{\setlength{\tabcolsep}{2pt}
\begin{tabular}[h]{@{}llllllrr@{}}
\toprule
\textbf{\# ~} & \textbf{Tool} & \textbf{Verified} & \textbf{Falsified} & \textbf{Fastest} & \textbf{Penalty} & \textbf{Score} & \textbf{Percent}\\
\midrule
1 & $\alpha$-$\beta$-CROWN & 8 & 13 & 0 & 0 & 210 & 100.0\% \\
2 & PyRAT & 8 & 11 & 0 & 0 & 190 & 90.5\% \\
3 & nnenum & 6 & 11 & 0 & 0 & 170 & 81.0\% \\
4 & NeuralSAT & 3 & 11 & 0 & 0 & 140 & 66.7\% \\
5 & NNV & 3 & 11 & 0 & 0 & 140 & 66.7\% \\
6 & Marabou & 0 & 13 & 0 & 0 & 130 & 61.9\% \\
\bottomrule
\end{tabular}
}
\end{center}
\end{table}

\begin{figure}[h]
\centerline{\includegraphics[width=\textwidth]{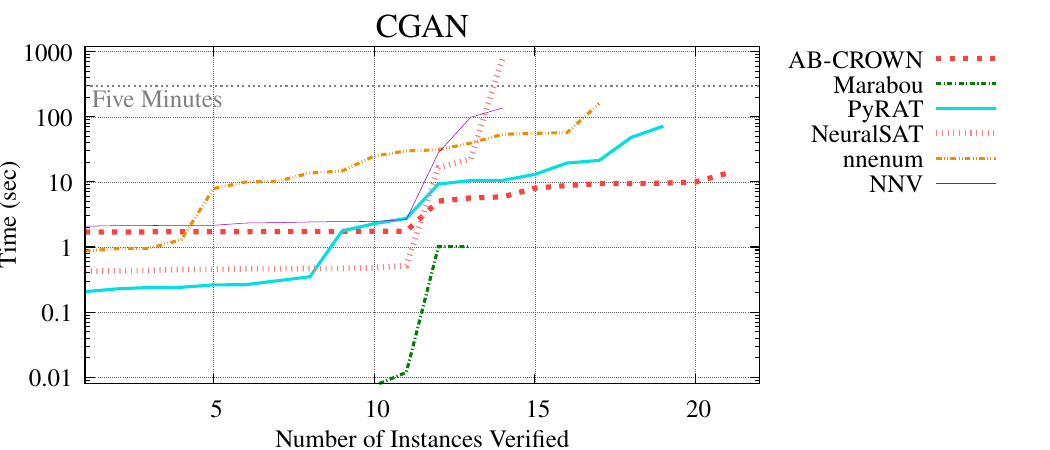}}
\caption{Cactus Plot for CGAN.}
\label{fig:quantPic}
\end{figure}


\begin{table}[h]
\begin{center}
\caption{Benchmark \texttt{2023-collins-rul-cnn}} \label{tab:cat_{cat}}
{\setlength{\tabcolsep}{2pt}
\begin{tabular}[h]{@{}llllllrr@{}}
\toprule
\textbf{\# ~} & \textbf{Tool} & \textbf{Verified} & \textbf{Falsified} & \textbf{Fastest} & \textbf{Penalty} & \textbf{Score} & \textbf{Percent}\\
\midrule
1 & nnenum & 35 & 27 & 0 & 0 & 620 & 100.0\% \\
2 & NeuralSAT & 35 & 27 & 0 & 0 & 620 & 100.0\% \\
3 & Marabou & 35 & 27 & 0 & 0 & 620 & 100.0\% \\
4 & $\alpha$-$\beta$-CROWN & 35 & 27 & 0 & 0 & 620 & 100.0\% \\
5 & PyRAT & 33 & 27 & 0 & 0 & 600 & 96.8\% \\
6 & NNV & 23 & 0 & 0 & 27 & -3820 & 0\% \\
\bottomrule
\end{tabular}
}
\end{center}
\end{table}

\begin{figure}[h]
\centerline{\includegraphics[width=\textwidth]{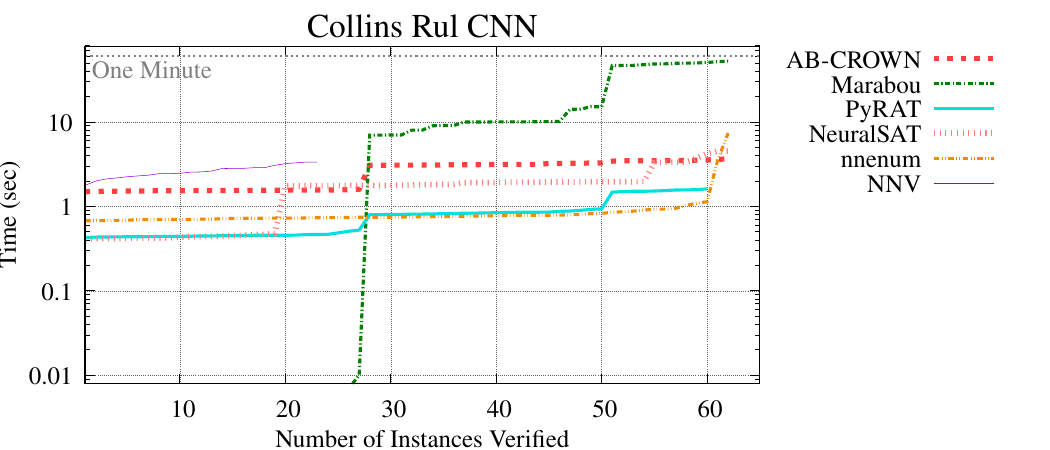}}
\caption{Cactus Plot for Collins Rul CNN.}
\label{fig:quantPic}
\end{figure}


\begin{table}[h]
\begin{center}
\caption{Benchmark \texttt{2023-dist-shift}} \label{tab:cat_{cat}}
{\setlength{\tabcolsep}{2pt}
\begin{tabular}[h]{@{}llllllrr@{}}
\toprule
\textbf{\# ~} & \textbf{Tool} & \textbf{Verified} & \textbf{Falsified} & \textbf{Fastest} & \textbf{Penalty} & \textbf{Score} & \textbf{Percent}\\
\midrule
1 & $\alpha$-$\beta$-CROWN & 67 & 5 & 0 & 0 & 720 & 100.0\% \\
2 & PyRAT & 66 & 5 & 0 & 0 & 710 & 98.6\% \\
3 & NeuralSAT & 66 & 5 & 0 & 0 & 710 & 98.6\% \\
4 & Marabou & 55 & 5 & 0 & 0 & 600 & 83.3\% \\
5 & NNV & 0 & 4 & 0 & 0 & 40 & 5.6\% \\
\bottomrule
\end{tabular}
}
\end{center}
\end{table}

\begin{figure}[h]
\centerline{\includegraphics[width=\textwidth]{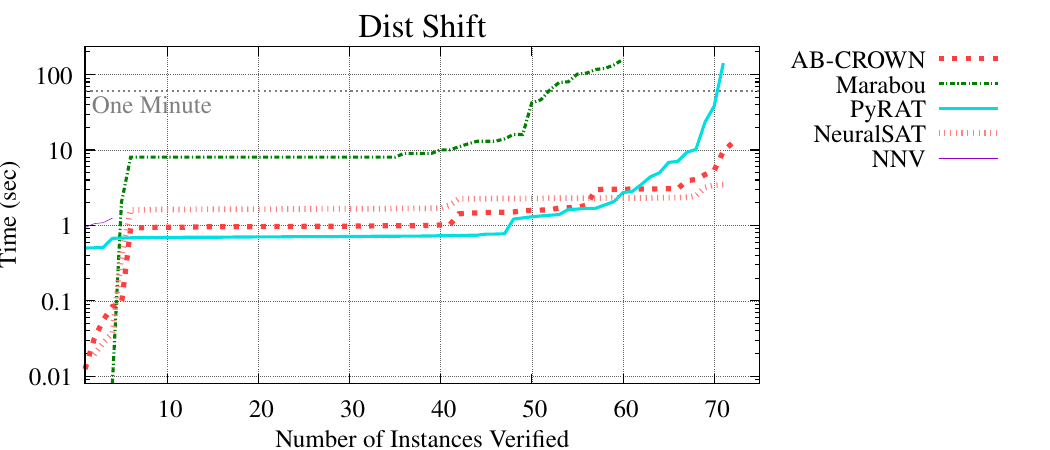}}
\caption{Cactus Plot for Dist Shift.}
\label{fig:quantPic}
\end{figure}


\begin{table}[h]
\begin{center}
\caption{Benchmark \texttt{2023-ml4acopf}} \label{tab:cat_{cat}}
{\setlength{\tabcolsep}{2pt}
\begin{tabular}[h]{@{}llllllrr@{}}
\toprule
\textbf{\# ~} & \textbf{Tool} & \textbf{Verified} & \textbf{Falsified} & \textbf{Fastest} & \textbf{Penalty} & \textbf{Score} & \textbf{Percent}\\
\midrule
1 & $\alpha$-$\beta$-CROWN & 18 & 1 & 0 & 0 & 190 & 100.0\% \\
\bottomrule
\end{tabular}
}
\end{center}
\end{table}

\begin{figure}[h]
\centerline{\includegraphics[width=\textwidth]{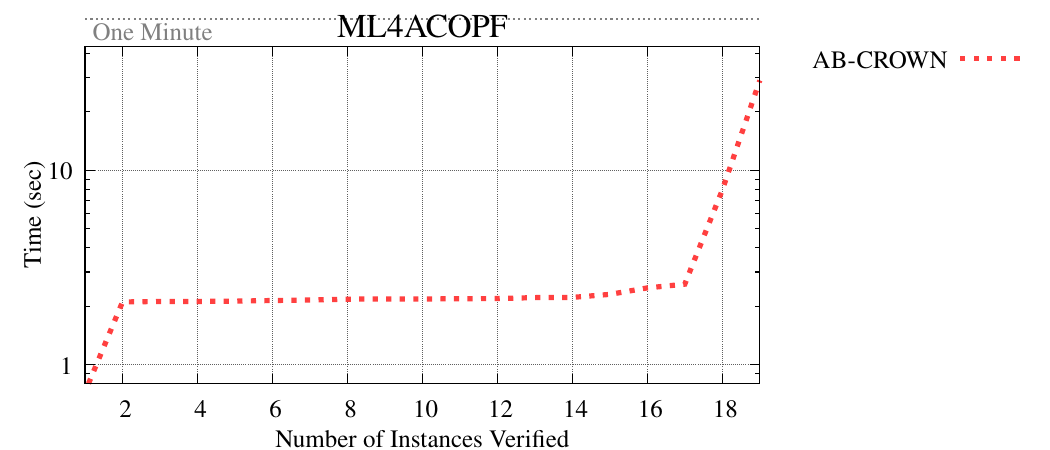}}
\caption{Cactus Plot for ML4ACOPF.}
\label{fig:quantPic}
\end{figure}


\begin{table}[h]
\begin{center}
\caption{Benchmark \texttt{2023-nn4sys}} \label{tab:cat_{cat}}
{\setlength{\tabcolsep}{2pt}
\begin{tabular}[h]{@{}llllllrr@{}}
\toprule
\textbf{\# ~} & \textbf{Tool} & \textbf{Verified} & \textbf{Falsified} & \textbf{Fastest} & \textbf{Penalty} & \textbf{Score} & \textbf{Percent}\\
\midrule
1 & $\alpha$-$\beta$-CROWN & 194 & 0 & 0 & 0 & 1940 & 100.0\% \\
2 & NeuralSAT & 83 & 0 & 0 & 0 & 830 & 42.8\% \\
3 & PyRAT & 40 & 0 & 0 & 0 & 400 & 20.6\% \\
4 & Marabou & 34 & 0 & 0 & 0 & 340 & 17.5\% \\
5 & nnenum & 22 & 0 & 0 & 0 & 220 & 11.3\% \\
6 & NNV & 0 & 2 & 0 & 8 & -1180 & 0\% \\
\bottomrule
\end{tabular}
}
\end{center}
\end{table}

\begin{figure}[h]
\centerline{\includegraphics[width=\textwidth]{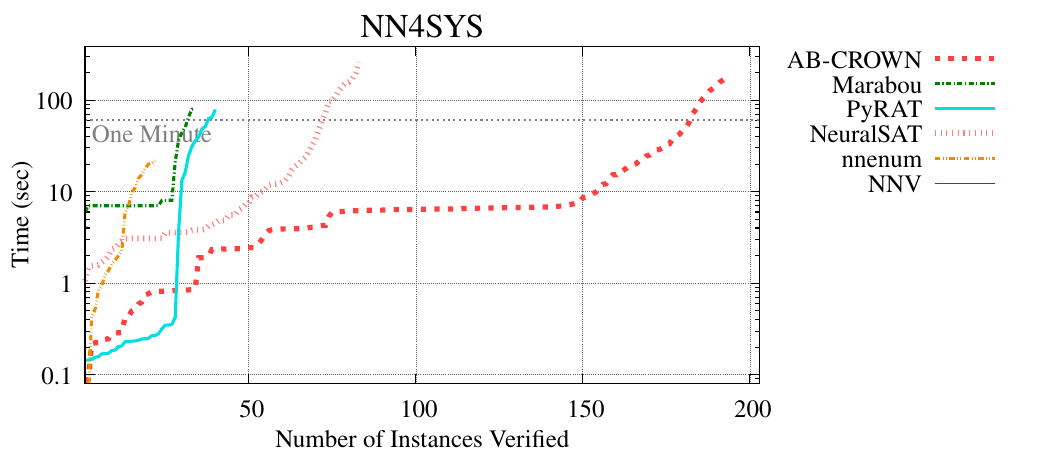}}
\caption{Cactus Plot for NN4SYS.}
\label{fig:quantPic}
\end{figure}


\begin{table}[h]
\begin{center}
\caption{Benchmark \texttt{2023-tllverifybench}} \label{tab:cat_{cat}}
{\setlength{\tabcolsep}{2pt}
\begin{tabular}[h]{@{}llllllrr@{}}
\toprule
\textbf{\# ~} & \textbf{Tool} & \textbf{Verified} & \textbf{Falsified} & \textbf{Fastest} & \textbf{Penalty} & \textbf{Score} & \textbf{Percent}\\
\midrule
1 & NeuralSAT & 15 & 17 & 0 & 0 & 320 & 100.0\% \\
2 & FastBATLLNN & 15 & 17 & 0 & 0 & 320 & 100.0\% \\
3 & $\alpha$-$\beta$-CROWN & 15 & 17 & 0 & 0 & 320 & 100.0\% \\
4 & PyRAT & 10 & 12 & 0 & 0 & 220 & 68.8\% \\
5 & Marabou & 5 & 17 & 0 & 0 & 220 & 68.8\% \\
6 & nnenum & 2 & 16 & 0 & 0 & 180 & 56.2\% \\
7 & NNV & 1 & 16 & 0 & 0 & 170 & 53.1\% \\
\bottomrule
\end{tabular}
}
\end{center}
\end{table}

\begin{figure}[h]
\centerline{\includegraphics[width=\textwidth]{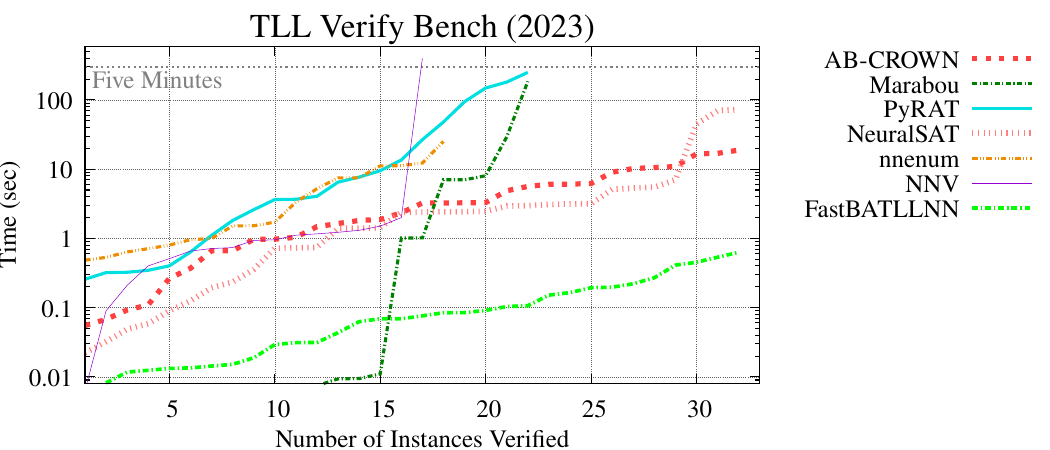}}
\caption{Cactus Plot for TLL Verify Bench (2023).}
\label{fig:quantPic}
\end{figure}


\begin{table}[h]
\begin{center}
\caption{Benchmark \texttt{2023-traffic-signs-recognition}} \label{tab:cat_{cat}}
{\setlength{\tabcolsep}{2pt}
\begin{tabular}[h]{@{}llllllrr@{}}
\toprule
\textbf{\# ~} & \textbf{Tool} & \textbf{Verified} & \textbf{Falsified} & \textbf{Fastest} & \textbf{Penalty} & \textbf{Score} & \textbf{Percent}\\
\midrule
1 & $\alpha$-$\beta$-CROWN & 0 & 42 & 0 & 0 & 420 & 100.0\% \\
2 & Marabou & 0 & 19 & 0 & 0 & 190 & 45.2\% \\
3 & PyRAT & 0 & 8 & 0 & 0 & 80 & 19.0\% \\
4 & NeuralSAT & 0 & 0 & 0 & 35 & -5250 & 0\% \\
\bottomrule
\end{tabular}
}
\end{center}
\end{table}

\begin{figure}[h]
\centerline{\includegraphics[width=\textwidth]{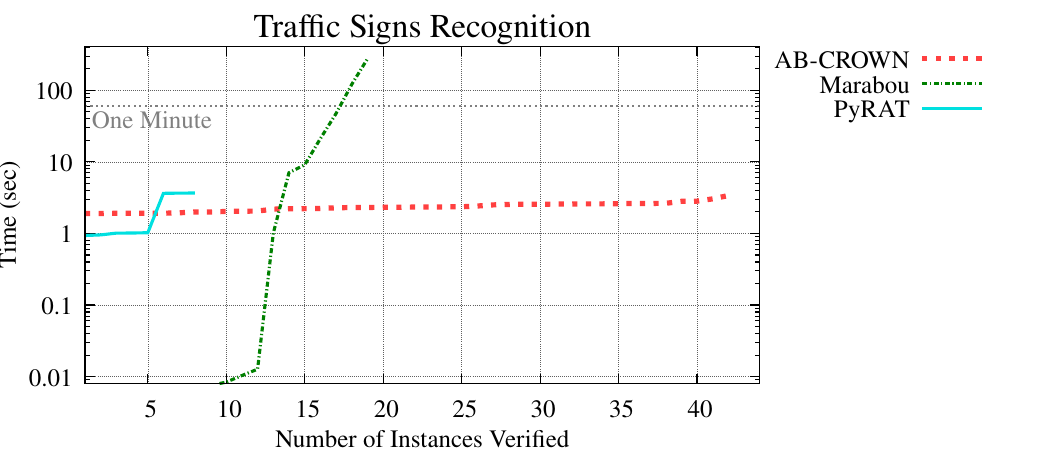}}
\caption{Cactus Plot for Traffic Signs Recognition.}
\label{fig:quantPic}
\end{figure}


\begin{table}[h]
\begin{center}
\caption{Benchmark \texttt{2023-vggnet16}} \label{tab:cat_{cat}}
{\setlength{\tabcolsep}{2pt}
\begin{tabular}[h]{@{}llllllrr@{}}
\toprule
\textbf{\# ~} & \textbf{Tool} & \textbf{Verified} & \textbf{Falsified} & \textbf{Fastest} & \textbf{Penalty} & \textbf{Score} & \textbf{Percent}\\
\midrule
1 & $\alpha$-$\beta$-CROWN & 15 & 0 & 0 & 0 & 150 & 100.0\% \\
2 & nnenum & 14 & 0 & 0 & 0 & 140 & 93.3\% \\
3 & PyRAT & 13 & 1 & 0 & 0 & 140 & 93.3\% \\
4 & NeuralSAT & 5 & 1 & 0 & 0 & 60 & 40.0\% \\
5 & Marabou & 2 & 1 & 0 & 0 & 30 & 20.0\% \\
\bottomrule
\end{tabular}
}
\end{center}
\end{table}

\begin{figure}[h]
\centerline{\includegraphics[width=\textwidth]{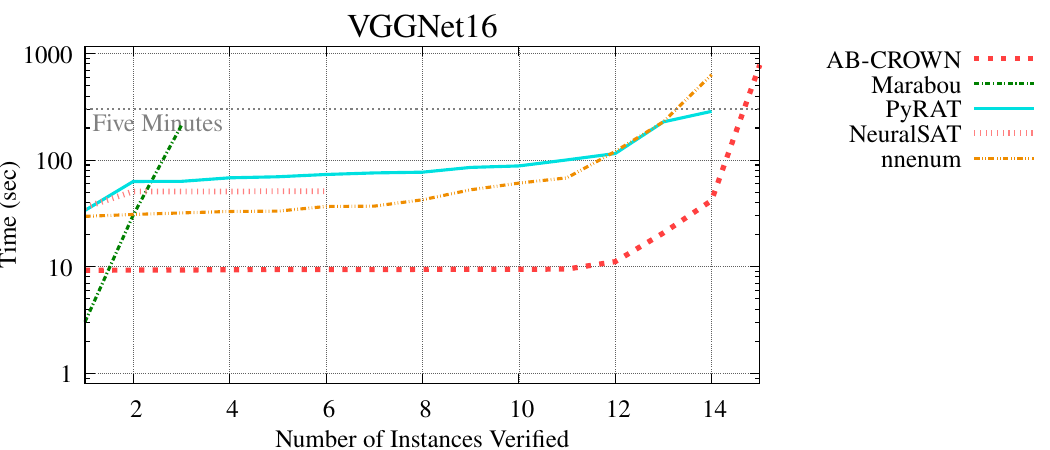}}
\caption{Cactus Plot for VGGNet16.}
\label{fig:quantPic}
\end{figure}


\begin{table}[h]
\begin{center}
\caption{Benchmark \texttt{2023-vit}. Note: A soundness issue was found after the competition in Marabou's results, due to an issue in Gurobi; these results may be updated in a future version of the report.} \label{tab:cat_{cat}}
{\setlength{\tabcolsep}{2pt}
\begin{tabular}[h]{@{}llllllrr@{}}
\toprule
\textbf{\# ~} & \textbf{Tool} & \textbf{Verified} & \textbf{Falsified} & \textbf{Fastest} & \textbf{Penalty} & \textbf{Score} & \textbf{Percent}\\
\midrule
1 & Marabou & 200 & 0 & 0 & 0 & 2000 & 100.0\% \\
2 & $\alpha$-$\beta$-CROWN & 79 & 0 & 0 & 0 & 790 & 39.5\% \\
\bottomrule
\end{tabular}
}
\end{center}
\end{table}

\begin{figure}[h]
\centerline{\includegraphics[width=\textwidth]{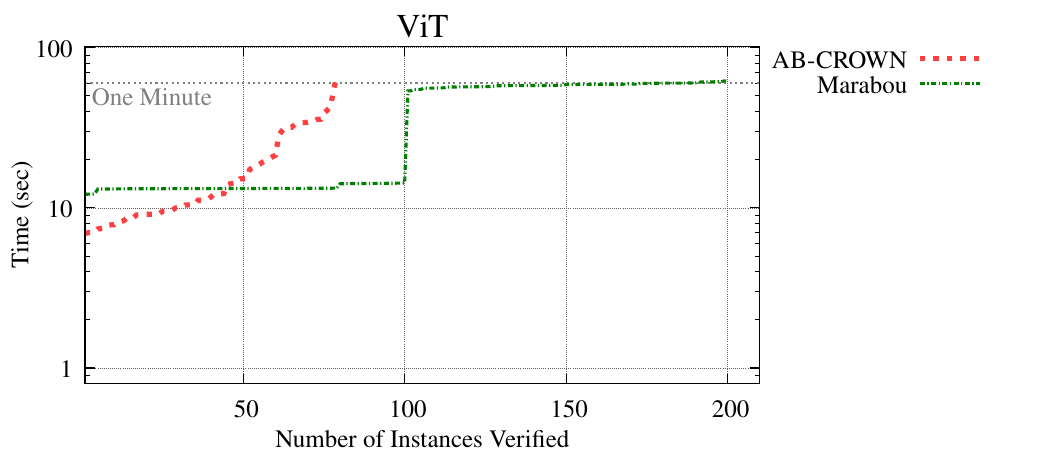}}
\caption{Cactus Plot for ViT.}
\label{fig:quantPic}
\end{figure}

\clearpage
\subsection{Unscored Benchmarks}
\label{sec:benchmark_results_unscored}


\begin{table}[h]
\begin{center}
\caption{Benchmark \texttt{2022-carvana-unet-2022}} \label{tab:cat_{cat}}
{\setlength{\tabcolsep}{2pt}
\begin{tabular}[h]{@{}llllllrr@{}}
\toprule
\textbf{\# ~} & \textbf{Tool} & \textbf{Verified} & \textbf{Falsified} & \textbf{Fastest} & \textbf{Penalty} & \textbf{Score} & \textbf{Percent}\\
\midrule
1 & $\alpha$-$\beta$-CROWN & 39 & 0 & 0 & 0 & 390 & 100.0\% \\
\bottomrule
\end{tabular}
}
\end{center}
\end{table}

\begin{figure}[h]
\centerline{\includegraphics[width=\textwidth]{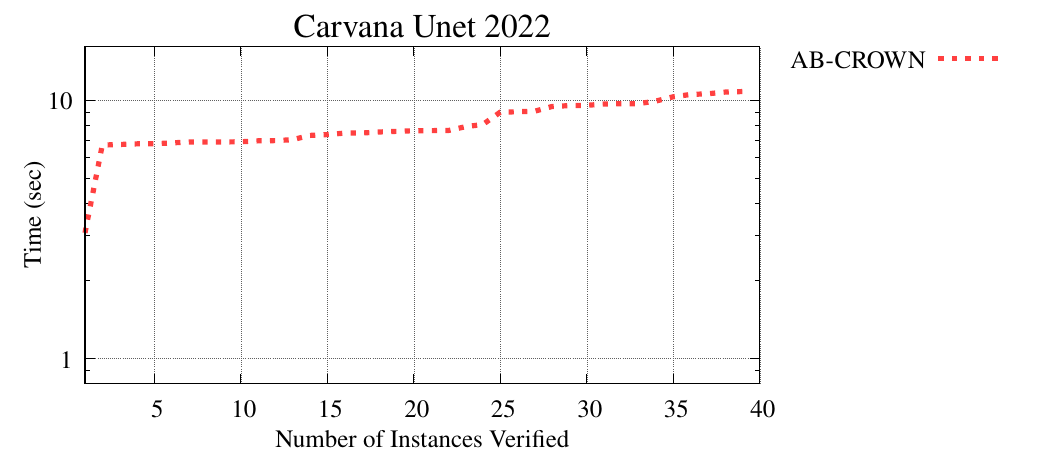}}
\caption{Cactus Plot for Carvana Unet 2022.}
\label{fig:quantPic}
\end{figure}


\begin{table}[h]
\begin{center}
\caption{Benchmark \texttt{2022-cifar100-tinyimagenet-resnet}} \label{tab:cat_{cat}}
{\setlength{\tabcolsep}{2pt}
\begin{tabular}[h]{@{}llllllrr@{}}
\toprule
\textbf{\# ~} & \textbf{Tool} & \textbf{Verified} & \textbf{Falsified} & \textbf{Fastest} & \textbf{Penalty} & \textbf{Score} & \textbf{Percent}\\
\midrule
1 & $\alpha$-$\beta$-CROWN & 69 & 3 & 0 & 0 & 720 & 100.0\% \\
2 & NeuralSAT & 56 & 3 & 0 & 0 & 590 & 81.9\% \\
\bottomrule
\end{tabular}
}
\end{center}
\end{table}

\begin{figure}[h]
\centerline{\includegraphics[width=\textwidth]{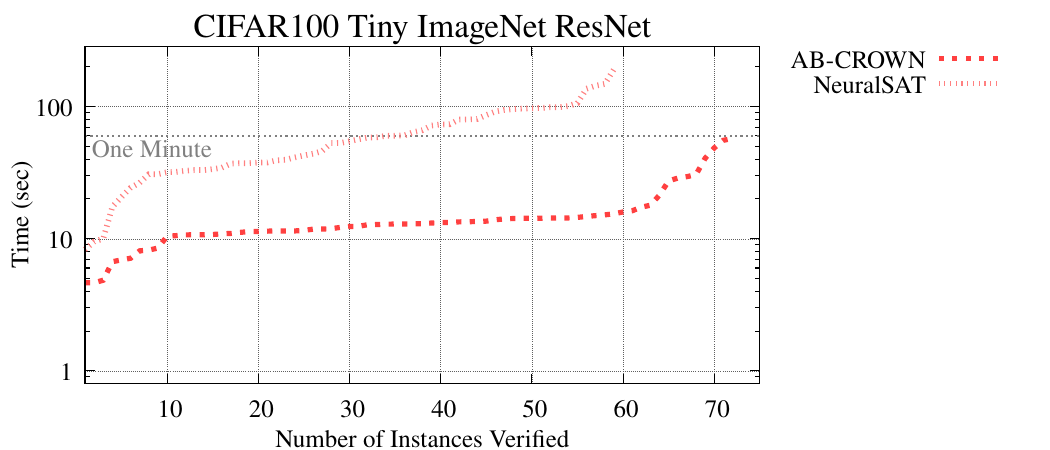}}
\caption{Cactus Plot for CIFAR100 Tiny ImageNet ResNet.}
\label{fig:quantPic}
\end{figure}


\begin{table}[h]
\begin{center}
\caption{Benchmark \texttt{2022-cifar2020}} \label{tab:cat_{cat}}
{\setlength{\tabcolsep}{2pt}
\begin{tabular}[h]{@{}llllllrr@{}}
\toprule
\textbf{\# ~} & \textbf{Tool} & \textbf{Verified} & \textbf{Falsified} & \textbf{Fastest} & \textbf{Penalty} & \textbf{Score} & \textbf{Percent}\\
\midrule
1 & NeuralSAT & 148 & 42 & 0 & 0 & 1900 & 100.0\% \\
2 & $\alpha$-$\beta$-CROWN & 148 & 41 & 0 & 0 & 1890 & 99.5\% \\
3 & PyRAT & 96 & 39 & 0 & 0 & 1350 & 71.1\% \\
4 & nnenum & 66 & 19 & 0 & 0 & 850 & 44.7\% \\
5 & NNV & 32 & 0 & 0 & 6 & -580 & 0\% \\
\bottomrule
\end{tabular}
}
\end{center}
\end{table}

\begin{figure}[h]
\centerline{\includegraphics[width=\textwidth]{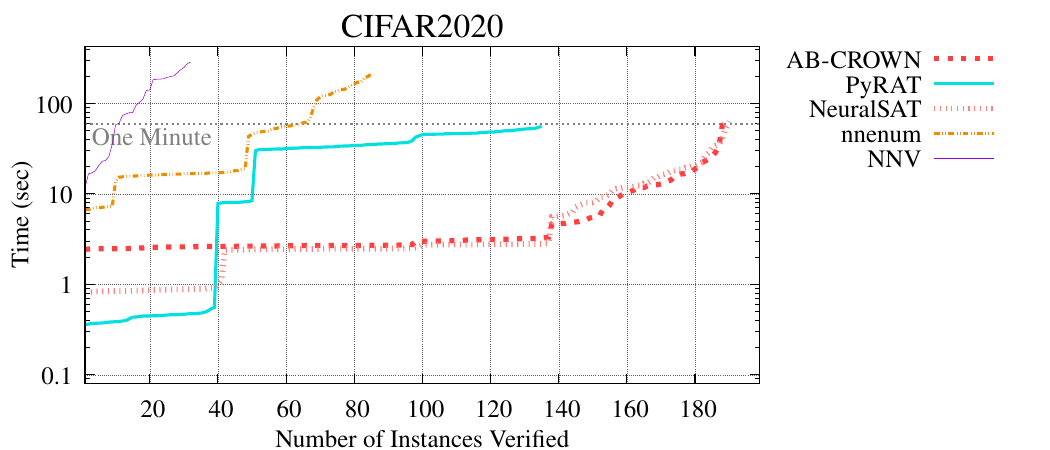}}
\caption{Cactus Plot for CIFAR2020.}
\label{fig:quantPic}
\end{figure}


\begin{table}[h]
\begin{center}
\caption{Benchmark \texttt{2022-cifar-biasfield}} \label{tab:cat_{cat}}
{\setlength{\tabcolsep}{2pt}
\begin{tabular}[h]{@{}llllllrr@{}}
\toprule
\textbf{\# ~} & \textbf{Tool} & \textbf{Verified} & \textbf{Falsified} & \textbf{Fastest} & \textbf{Penalty} & \textbf{Score} & \textbf{Percent}\\
\midrule
1 & $\alpha$-$\beta$-CROWN & 68 & 1 & 0 & 0 & 690 & 100.0\% \\
2 & NeuralSAT & 67 & 1 & 0 & 0 & 680 & 98.6\% \\
3 & PyRAT & 51 & 1 & 0 & 0 & 520 & 75.4\% \\
4 & nnenum & 18 & 0 & 0 & 0 & 180 & 26.1\% \\
\bottomrule
\end{tabular}
}
\end{center}
\end{table}

\begin{figure}[h]
\centerline{\includegraphics[width=\textwidth]{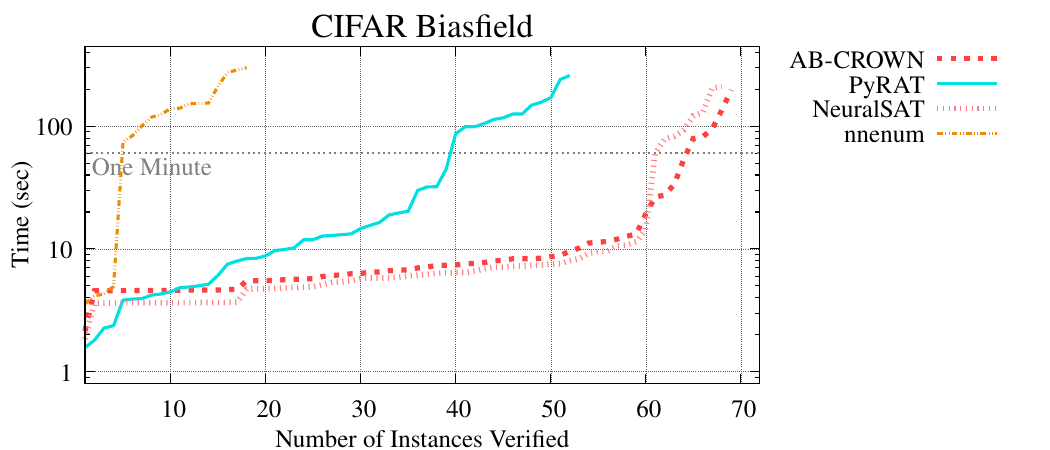}}
\caption{Cactus Plot for CIFAR Biasfield.}
\label{fig:quantPic}
\end{figure}


\begin{table}[h]
\begin{center}
\caption{Benchmark \texttt{2022-mnist-fc}} \label{tab:cat_{cat}}
{\setlength{\tabcolsep}{2pt}
\begin{tabular}[h]{@{}llllllrr@{}}
\toprule
\textbf{\# ~} & \textbf{Tool} & \textbf{Verified} & \textbf{Falsified} & \textbf{Fastest} & \textbf{Penalty} & \textbf{Score} & \textbf{Percent}\\
\midrule
1 & NeuralSAT & 46 & 18 & 0 & 0 & 640 & 100.0\% \\
2 & nnenum & 48 & 11 & 0 & 0 & 590 & 92.2\% \\
3 & $\alpha$-$\beta$-CROWN & 68 & 16 & 0 & 2 & 540 & 84.4\% \\
4 & PyRAT & 35 & 16 & 0 & 0 & 510 & 79.7\% \\
5 & NNV & 30 & 4 & 0 & 0 & 340 & 53.1\% \\
\bottomrule
\end{tabular}
}
\end{center}
\end{table}

\begin{figure}[h]
\centerline{\includegraphics[width=\textwidth]{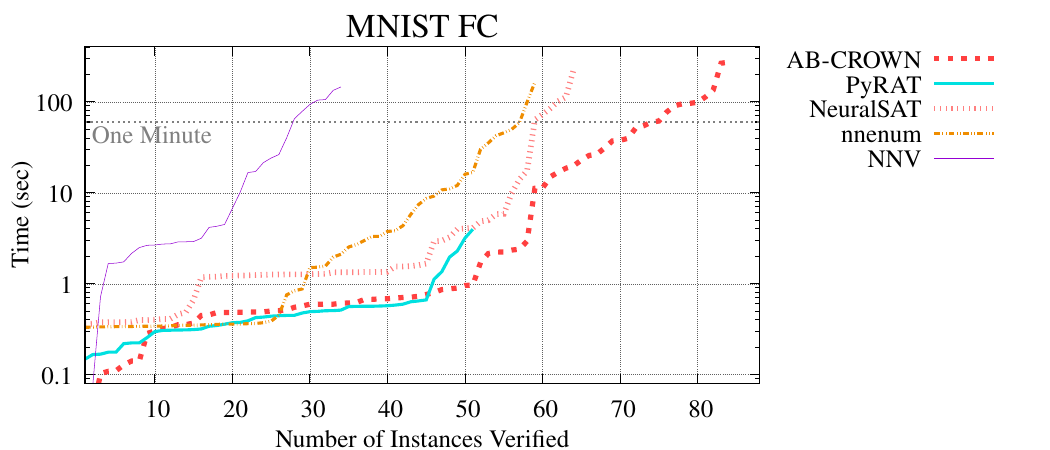}}
\caption{Cactus Plot for MNIST FC.}
\label{fig:quantPic}
\end{figure}


\begin{table}[h]
\begin{center}
\caption{Benchmark \texttt{2022-nn4sys}} \label{tab:cat_{cat}}
{\setlength{\tabcolsep}{2pt}
\begin{tabular}[h]{@{}llllllrr@{}}
\toprule
\textbf{\# ~} & \textbf{Tool} & \textbf{Verified} & \textbf{Falsified} & \textbf{Fastest} & \textbf{Penalty} & \textbf{Score} & \textbf{Percent}\\
\midrule
1 & $\alpha$-$\beta$-CROWN & 152 & 0 & 0 & 0 & 1520 & 100.0\% \\
2 & NeuralSAT & 78 & 0 & 0 & 0 & 780 & 51.3\% \\
3 & nnenum & 22 & 0 & 0 & 0 & 220 & 14.5\% \\
4 & PyRAT & 14 & 0 & 0 & 0 & 140 & 9.2\% \\
5 & NNV & 0 & 2 & 0 & 8 & -1180 & 0\% \\
\bottomrule
\end{tabular}
}
\end{center}
\end{table}

\begin{figure}[h]
\centerline{\includegraphics[width=\textwidth]{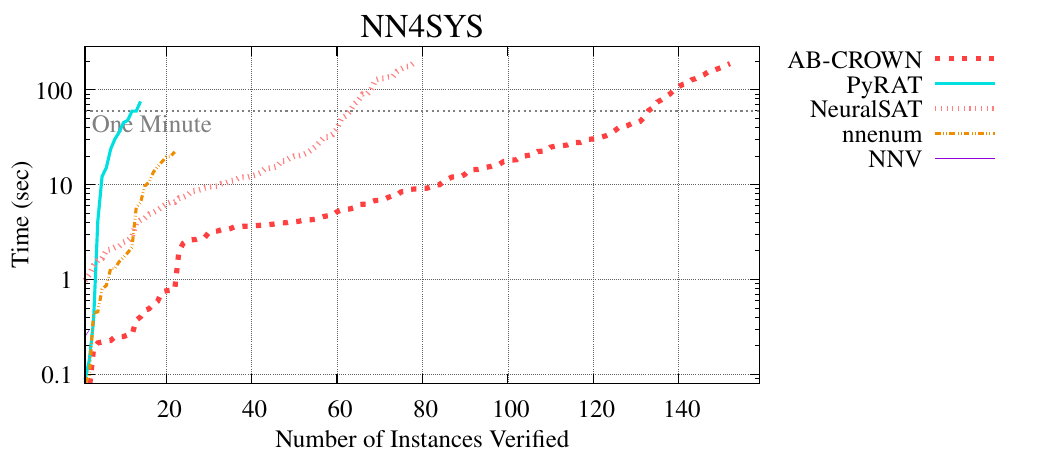}}
\caption{Cactus Plot for NN4SYS.}
\label{fig:quantPic}
\end{figure}


\begin{table}[h]
\begin{center}
\caption{Benchmark \texttt{2022-oval21}} \label{tab:cat_{cat}}
{\setlength{\tabcolsep}{2pt}
\begin{tabular}[h]{@{}llllllrr@{}}
\toprule
\textbf{\# ~} & \textbf{Tool} & \textbf{Verified} & \textbf{Falsified} & \textbf{Fastest} & \textbf{Penalty} & \textbf{Score} & \textbf{Percent}\\
\midrule
1 & $\alpha$-$\beta$-CROWN & 25 & 1 & 0 & 0 & 260 & 100.0\% \\
2 & NeuralSAT & 19 & 1 & 0 & 0 & 200 & 76.9\% \\
3 & nnenum & 3 & 1 & 0 & 0 & 40 & 15.4\% \\
4 & NNV & 4 & 0 & 0 & 0 & 40 & 15.4\% \\
5 & PyRAT & 1 & 1 & 0 & 0 & 20 & 7.7\% \\
\bottomrule
\end{tabular}
}
\end{center}
\end{table}

\begin{figure}[h]
\centerline{\includegraphics[width=\textwidth]{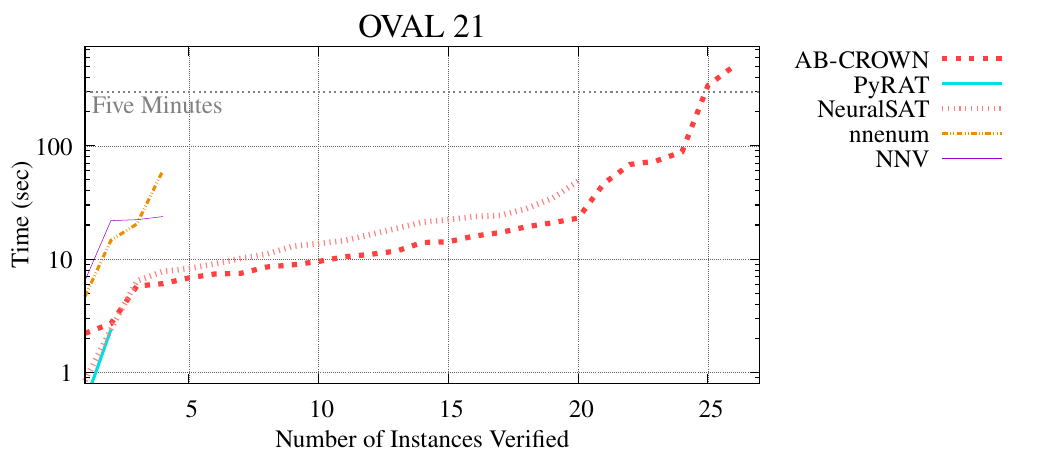}}
\caption{Cactus Plot for OVAL 21.}
\label{fig:quantPic}
\end{figure}


\begin{table}[h]
\begin{center}
\caption{Benchmark \texttt{2022-reach-prob-density}} \label{tab:cat_{cat}}
{\setlength{\tabcolsep}{2pt}
\begin{tabular}[h]{@{}llllllrr@{}}
\toprule
\textbf{\# ~} & \textbf{Tool} & \textbf{Verified} & \textbf{Falsified} & \textbf{Fastest} & \textbf{Penalty} & \textbf{Score} & \textbf{Percent}\\
\midrule
1 & nnenum & 22 & 14 & 0 & 0 & 360 & 100.0\% \\
2 & $\alpha$-$\beta$-CROWN & 22 & 14 & 0 & 0 & 360 & 100.0\% \\
3 & NeuralSAT & 22 & 12 & 0 & 0 & 340 & 94.4\% \\
4 & PyRAT & 17 & 7 & 0 & 0 & 240 & 66.7\% \\
5 & NNV & 9 & 0 & 0 & 0 & 90 & 25.0\% \\
\bottomrule
\end{tabular}
}
\end{center}
\end{table}

\begin{figure}[h]
\centerline{\includegraphics[width=\textwidth]{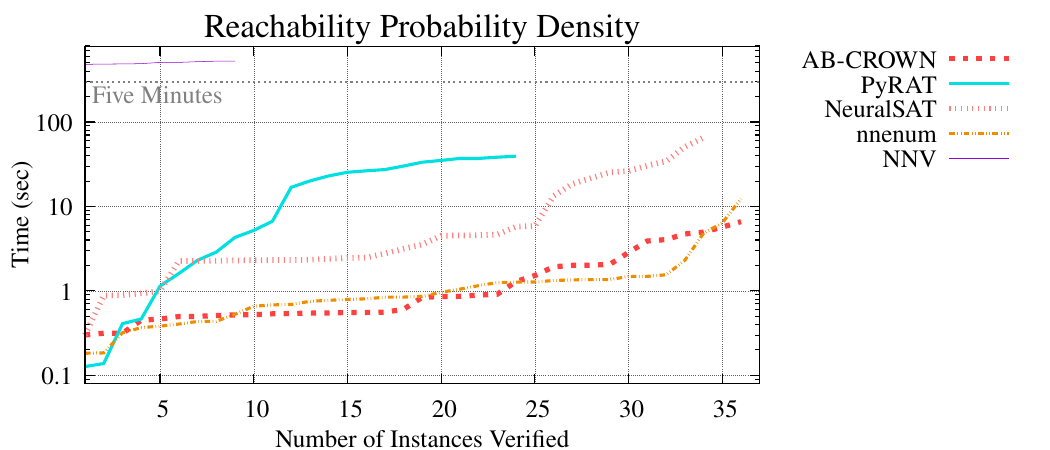}}
\caption{Cactus Plot for Reachability Probability Density.}
\label{fig:quantPic}
\end{figure}


\begin{table}[h]
\begin{center}
\caption{Benchmark \texttt{2022-rl-benchmarks}} \label{tab:cat_{cat}}
{\setlength{\tabcolsep}{2pt}
\begin{tabular}[h]{@{}llllllrr@{}}
\toprule
\textbf{\# ~} & \textbf{Tool} & \textbf{Verified} & \textbf{Falsified} & \textbf{Fastest} & \textbf{Penalty} & \textbf{Score} & \textbf{Percent}\\
\midrule
1 & NeuralSAT & 193 & 103 & 0 & 0 & 2960 & 100.0\% \\
2 & $\alpha$-$\beta$-CROWN & 193 & 103 & 0 & 0 & 2960 & 100.0\% \\
3 & nnenum & 191 & 103 & 0 & 0 & 2940 & 99.3\% \\
4 & PyRAT & 194 & 96 & 0 & 0 & 2900 & 98.0\% \\
5 & NNV & 177 & 90 & 0 & 0 & 2670 & 90.2\% \\
\bottomrule
\end{tabular}
}
\end{center}
\end{table}

\begin{figure}[h]
\centerline{\includegraphics[width=\textwidth]{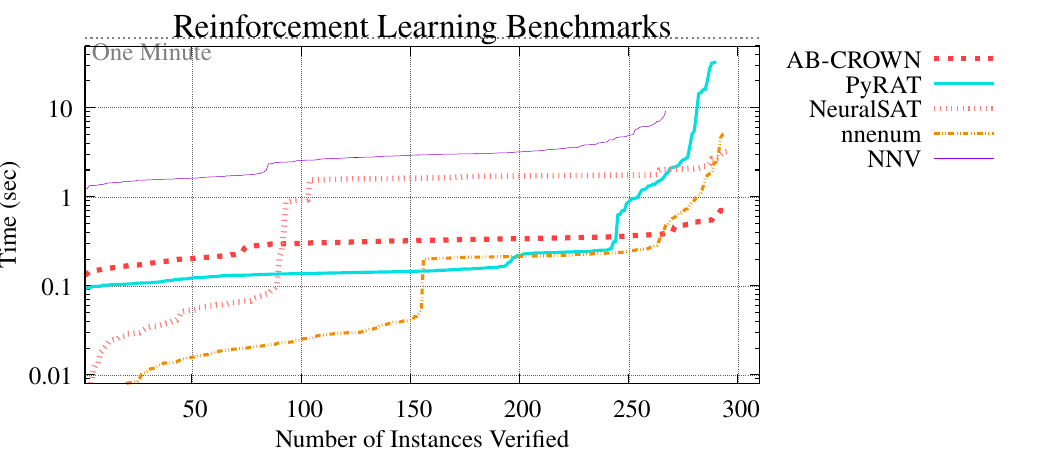}}
\caption{Cactus Plot for Reinforcement Learning Benchmarks.}
\label{fig:quantPic}
\end{figure}


\begin{table}[h]
\begin{center}
\caption{Benchmark \texttt{2022-sri-resnet-a}} \label{tab:cat_{cat}}
{\setlength{\tabcolsep}{2pt}
\begin{tabular}[h]{@{}llllllrr@{}}
\toprule
\textbf{\# ~} & \textbf{Tool} & \textbf{Verified} & \textbf{Falsified} & \textbf{Fastest} & \textbf{Penalty} & \textbf{Score} & \textbf{Percent}\\
\midrule
1 & $\alpha$-$\beta$-CROWN & 20 & 12 & 0 & 0 & 320 & 100.0\% \\
2 & NeuralSAT & 18 & 12 & 0 & 0 & 300 & 93.8\% \\
3 & PyRAT & 5 & 11 & 0 & 0 & 160 & 50.0\% \\
\bottomrule
\end{tabular}
}
\end{center}
\end{table}

\begin{figure}[h]
\centerline{\includegraphics[width=\textwidth]{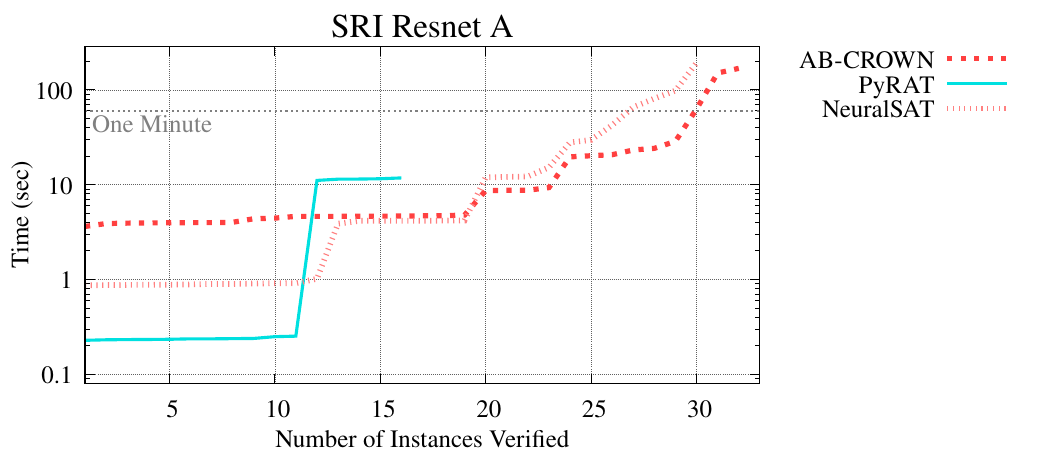}}
\caption{Cactus Plot for SRI Resnet A.}
\label{fig:quantPic}
\end{figure}


\begin{table}[h]
\begin{center}
\caption{Benchmark \texttt{2022-sri-resnet-b}} \label{tab:cat_{cat}}
{\setlength{\tabcolsep}{2pt}
\begin{tabular}[h]{@{}llllllrr@{}}
\toprule
\textbf{\# ~} & \textbf{Tool} & \textbf{Verified} & \textbf{Falsified} & \textbf{Fastest} & \textbf{Penalty} & \textbf{Score} & \textbf{Percent}\\
\midrule
1 & NeuralSAT & 28 & 11 & 0 & 0 & 390 & 100.0\% \\
2 & $\alpha$-$\beta$-CROWN & 28 & 11 & 0 & 0 & 390 & 100.0\% \\
3 & PyRAT & 13 & 11 & 0 & 0 & 240 & 61.5\% \\
4 & NNV & 2 & 0 & 0 & 0 & 20 & 5.1\% \\
\bottomrule
\end{tabular}
}
\end{center}
\end{table}

\begin{figure}[h]
\centerline{\includegraphics[width=\textwidth]{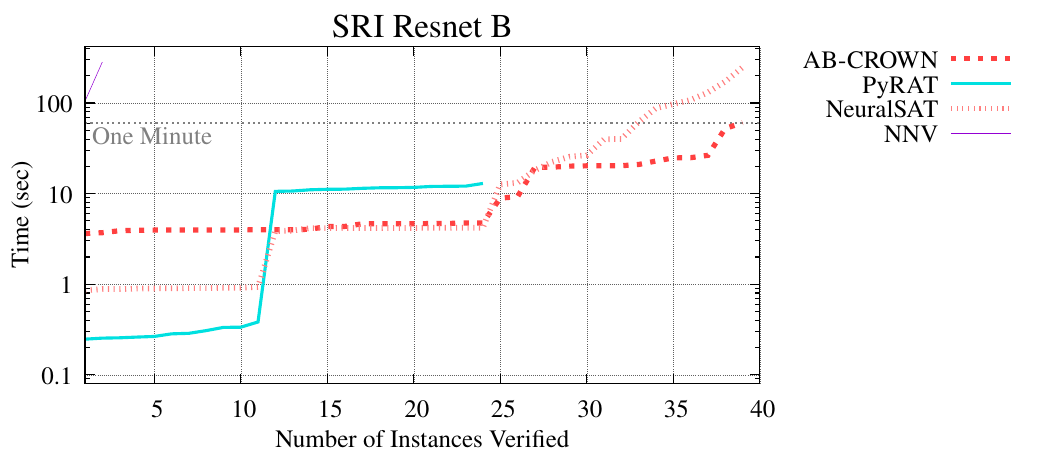}}
\caption{Cactus Plot for SRI Resnet B.}
\label{fig:quantPic}
\end{figure}


\begin{table}[h]
\begin{center}
\caption{Benchmark \texttt{2022-tllverifybench}} \label{tab:cat_{cat}}
{\setlength{\tabcolsep}{2pt}
\begin{tabular}[h]{@{}llllllrr@{}}
\toprule
\textbf{\# ~} & \textbf{Tool} & \textbf{Verified} & \textbf{Falsified} & \textbf{Fastest} & \textbf{Penalty} & \textbf{Score} & \textbf{Percent}\\
\midrule
1 & NeuralSAT & 11 & 21 & 0 & 0 & 320 & 100.0\% \\
2 & FastBATLLNN & 11 & 21 & 0 & 0 & 320 & 100.0\% \\
3 & $\alpha$-$\beta$-CROWN & 11 & 21 & 0 & 0 & 320 & 100.0\% \\
4 & PyRAT & 9 & 17 & 0 & 0 & 260 & 81.2\% \\
5 & nnenum & 1 & 21 & 0 & 0 & 220 & 68.8\% \\
6 & NNV & 0 & 21 & 0 & 0 & 210 & 65.6\% \\
\bottomrule
\end{tabular}
}
\end{center}
\end{table}

\begin{figure}[h]
\centerline{\includegraphics[width=\textwidth]{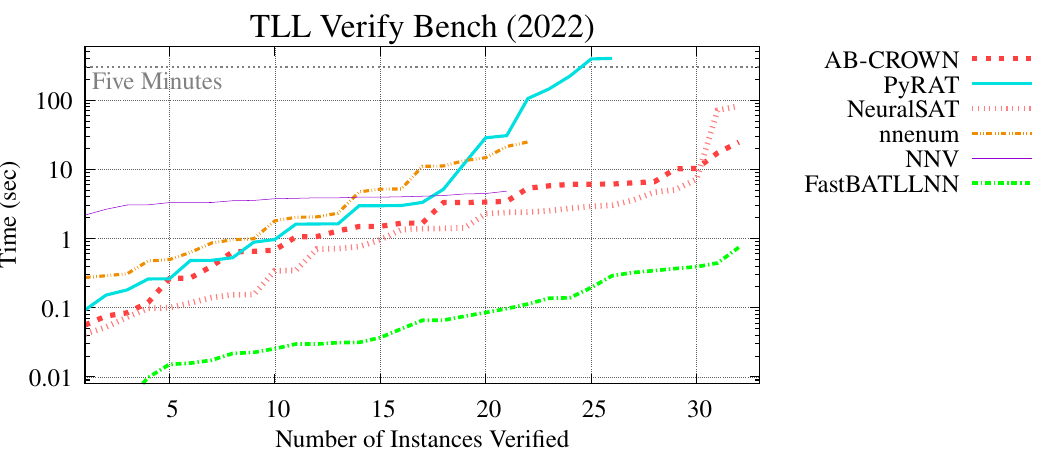}}
\caption{Cactus Plot for TLL Verify Bench (2022).}
\label{fig:quantPic}
\end{figure}


\begin{table}[h]
\begin{center}
\caption{Benchmark \texttt{2022-vggnet16-2022}} \label{tab:cat_{cat}}
{\setlength{\tabcolsep}{2pt}
\begin{tabular}[h]{@{}llllllrr@{}}
\toprule
\textbf{\# ~} & \textbf{Tool} & \textbf{Verified} & \textbf{Falsified} & \textbf{Fastest} & \textbf{Penalty} & \textbf{Score} & \textbf{Percent}\\
\midrule
1 & nnenum & 11 & 1 & 0 & 0 & 120 & 100.0\% \\
2 & PyRAT & 9 & 1 & 0 & 0 & 100 & 83.3\% \\
3 & NeuralSAT & 5 & 1 & 0 & 0 & 60 & 50.0\% \\
4 & $\alpha$-$\beta$-CROWN & 14 & 0 & 0 & 1 & -10 & 0\% \\
\bottomrule
\end{tabular}
}
\end{center}
\end{table}

\begin{figure}[h]
\centerline{\includegraphics[width=\textwidth]{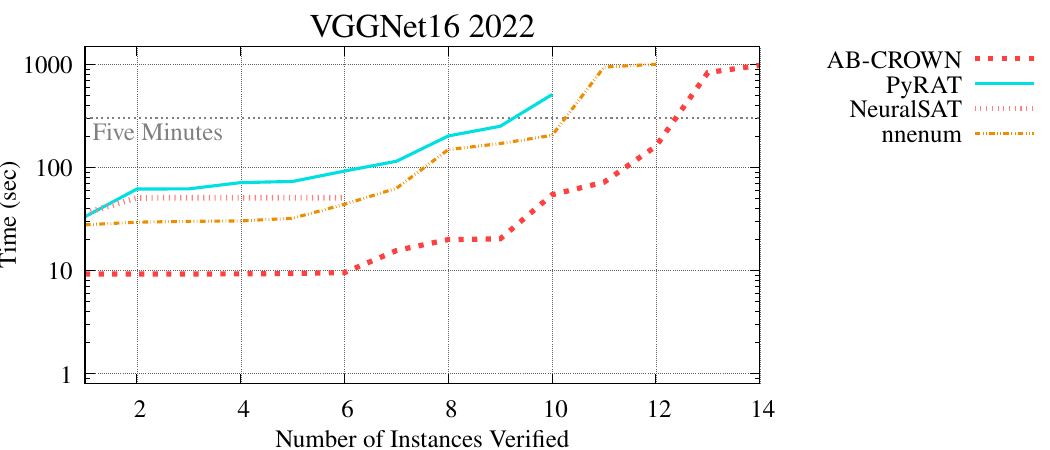}}
\caption{Cactus Plot for VGGNet16 2022.}
\label{fig:quantPic}
\end{figure}


\begin{table}[h]
\begin{center}
\caption{Benchmark \texttt{2023-cctsdb-yolo}} \label{tab:cat_{cat}}
{\setlength{\tabcolsep}{2pt}
\begin{tabular}[h]{@{}llllllrr@{}}
\toprule
\textbf{\# ~} & \textbf{Tool} & \textbf{Verified} & \textbf{Falsified} & \textbf{Fastest} & \textbf{Penalty} & \textbf{Score} & \textbf{Percent}\\
\midrule
1 & $\alpha$-$\beta$-CROWN & 11 & 28 & 0 & 0 & 390 & 100.0\% \\
\bottomrule
\end{tabular}
}
\end{center}
\end{table}

\begin{figure}[h]
\centerline{\includegraphics[width=\textwidth]{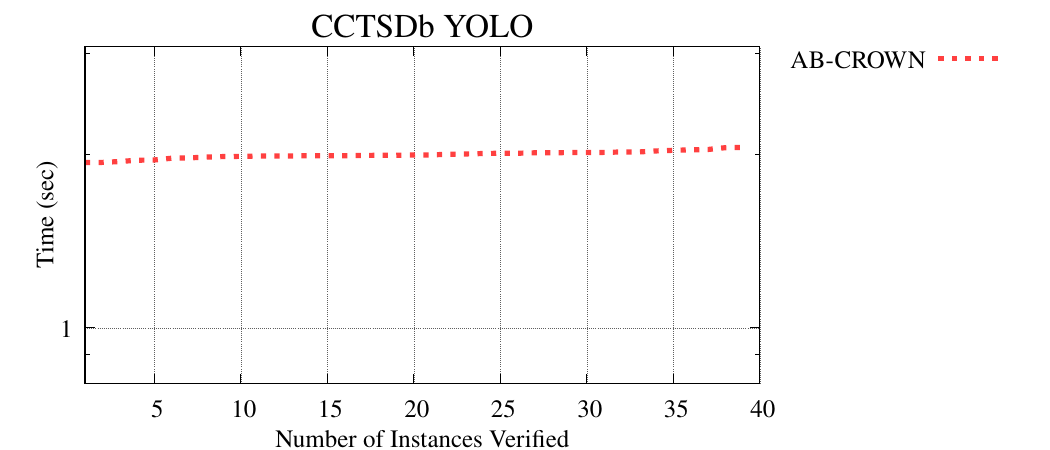}}
\caption{Cactus Plot for CCTSDb YOLO.}
\label{fig:quantPic}
\end{figure}


\begin{table}[h]
\begin{center}
\caption{Benchmark \texttt{2023-collins-yolo-robustness}} \label{tab:cat_{cat}}
{\setlength{\tabcolsep}{2pt}
\begin{tabular}[h]{@{}llllllrr@{}}
\toprule
\textbf{\# ~} & \textbf{Tool} & \textbf{Verified} & \textbf{Falsified} & \textbf{Fastest} & \textbf{Penalty} & \textbf{Score} & \textbf{Percent}\\
\midrule
1 & PyRAT & 0 & 6 & 0 & 0 & 60 & 100.0\% \\
2 & $\alpha$-$\beta$-CROWN & 0 & 6 & 0 & 0 & 60 & 100.0\% \\
\bottomrule
\end{tabular}
}
\end{center}
\end{table}

\begin{figure}[h]
\centerline{\includegraphics[width=\textwidth]{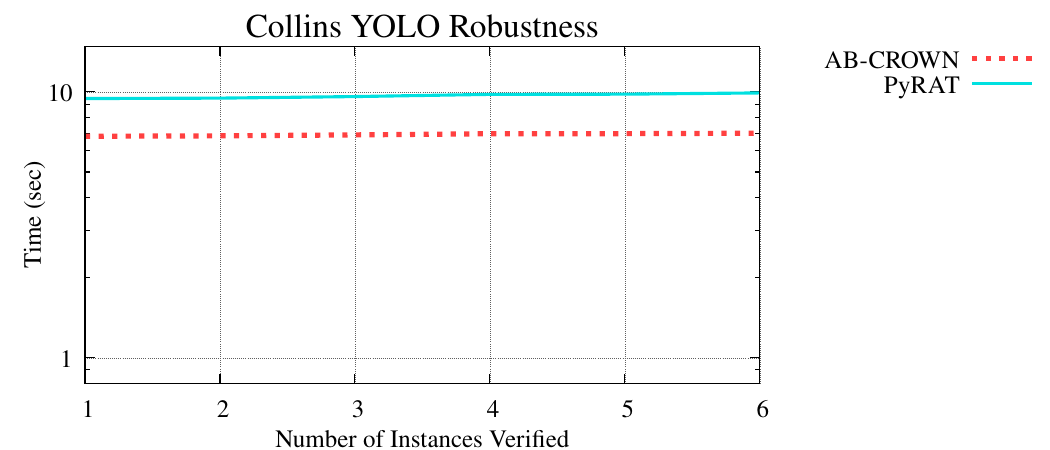}}
\caption{Cactus Plot for Collins YOLO Robustness.}
\label{fig:quantPic}
\end{figure}


\begin{table}[h]
\begin{center}
\caption{Benchmark \texttt{2023-metaroom}} \label{tab:cat_{cat}}
{\setlength{\tabcolsep}{2pt}
\begin{tabular}[h]{@{}llllllrr@{}}
\toprule
\textbf{\# ~} & \textbf{Tool} & \textbf{Verified} & \textbf{Falsified} & \textbf{Fastest} & \textbf{Penalty} & \textbf{Score} & \textbf{Percent}\\
\midrule
1 & $\alpha$-$\beta$-CROWN & 94 & 5 & 0 & 0 & 990 & 100.0\% \\
2 & NeuralSAT & 92 & 5 & 0 & 0 & 970 & 98.0\% \\
3 & PyRAT & 93 & 2 & 0 & 0 & 950 & 96.0\% \\
4 & nnenum & 51 & 2 & 0 & 0 & 530 & 53.5\% \\
\bottomrule
\end{tabular}
}
\end{center}
\end{table}

\begin{figure}[h]
\centerline{\includegraphics[width=\textwidth]{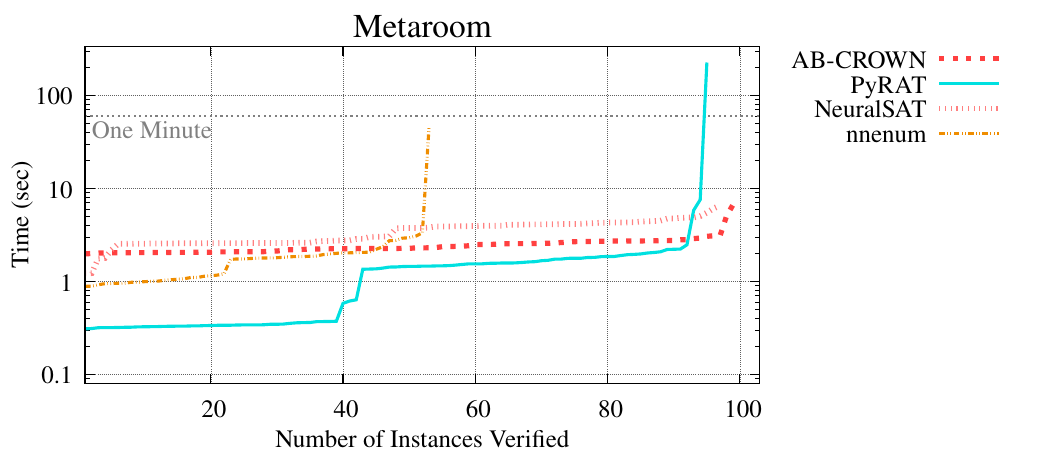}}
\caption{Cactus Plot for Metaroom.}
\label{fig:quantPic}
\end{figure}


\begin{table}[h]
\begin{center}
\caption{Benchmark \texttt{2023-yolo}} \label{tab:cat_{cat}}
{\setlength{\tabcolsep}{2pt}
\begin{tabular}[h]{@{}llllllrr@{}}
\toprule
\textbf{\# ~} & \textbf{Tool} & \textbf{Verified} & \textbf{Falsified} & \textbf{Fastest} & \textbf{Penalty} & \textbf{Score} & \textbf{Percent}\\
\midrule
1 & $\alpha$-$\beta$-CROWN & 63 & 0 & 0 & 0 & 630 & 100.0\% \\
2 & PyRAT & 41 & 0 & 0 & 0 & 410 & 65.1\% \\
3 & NNV & 0 & 0 & 0 & 41 & -6150 & 0\% \\
\bottomrule
\end{tabular}
}
\end{center}
\end{table}

\begin{figure}[h]
\centerline{\includegraphics[width=\textwidth]{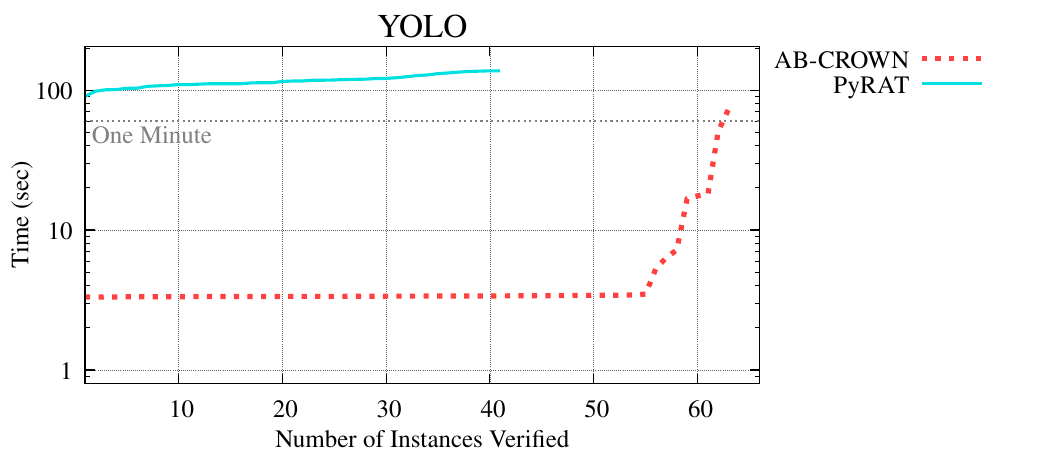}}
\caption{Cactus Plot for YOLO.}
\label{fig:quantPic}
\end{figure}

\clearpage
\section{Alternative Scoring Results}
\label{sec:alternative_ranking}
As described in Section~\ref{sec:scoring}, tools had to provide a concrete counterexample if they report SAT.
These were supposed to contain both the network input and the corresponding output.
However, as several tools provided either no outputs or incorrect outputs, we decided to discard all outputs in the provided counterexamples.
Instead, the outputs were computed using the onnxruntime package.

Alternatively, tools could have been assigned a penalty if the network output for the counterexample was missing or incorrect. 
This would have lead to the following ranking:

\begin{table}[h]
\begin{center}
\caption{Overall Score} \label{tab:score}
{\setlength{\tabcolsep}{2pt}
\begin{tabular}[h]{@{}lll@{}}
\toprule
\textbf{\# ~} & \textbf{Tool} & \textbf{Score}\\
\midrule
1 & $\alpha$-$\beta$-CROWN & 830.9 \\
2 & Marabou & 531.6 \\
3 & NeuralSAT & 469.6 \\
4 & nnenum & 441.9 \\
5 & PyRAT & 276.6 \\
6 & NNV & 176.4 \\
7 & FastBATLLNN & 100.0 \\
\bottomrule
\end{tabular}
}
\end{center}
\end{table}

The detailed list of results for this setting can be generated using the scripts in \url{https://github.com/ChristopherBrix/vnncomp2023_results}.

\clearpage
\subsection{Detailed Results}
\label{sec:results_detailed}

\begin{center}
{\setlength{\tabcolsep}{1pt}
\scriptsize
\begin{longtable}{@{}llllllllll@{}}
\caption{\footnotesize Instance Runtimes. Fastest times are \textcolor{blue}{blue}. Second fastest are \textcolor{second}{green}. Penalties are red crosses (\textbf{\textcolor{red}{\ding{55}}}).} \label{tab:all_results} \\
\toprule
\textbf{Category} & \textbf{Id} & \textbf{Result} & \textbf{$\alpha$-$\beta$-C} & \textbf{Marab} & \textbf{PyRat} & \textbf{NSAT} & \textbf{NNen} & \textbf{NNV} & \textbf{FastBaT} \\
\midrule
\endhead
2023 Acasxu & 0 & ~\textsc{unsat} & \textcolor{darkgray}{2.47} & \textcolor{darkgray}{7.02} & \textcolor{darkgray}{1.39} & \textcolor{darkgray}{3.23} & \textcolor{darkgray}{0.43} & - & - \\
2023 Acasxu & 1 & ~\textsc{unsat} & \textcolor{darkgray}{2.40} & \textcolor{darkgray}{7.02} & \textcolor{darkgray}{1.60} & \textcolor{darkgray}{3.16} & \textcolor{darkgray}{0.42} & - & - \\
2023 Acasxu & 2 & ~\textsc{unsat} & \textcolor{darkgray}{2.46} & \textcolor{darkgray}{7.02} & \textcolor{darkgray}{2.32} & \textcolor{darkgray}{3.23} & \textcolor{darkgray}{0.60} & - & - \\
2023 Acasxu & 3 & ~\textsc{unsat} & \textcolor{darkgray}{2.45} & \textcolor{darkgray}{7.02} & \textcolor{darkgray}{1.82} & \textcolor{darkgray}{3.16} & \textcolor{darkgray}{0.62} & - & - \\
2023 Acasxu & 4 & ~\textsc{unsat} & \textcolor{darkgray}{2.36} & \textcolor{darkgray}{7.02} & \textcolor{darkgray}{1.29} & \textcolor{darkgray}{3.15} & \textcolor{darkgray}{0.49} & - & - \\
2023 Acasxu & 5 & ~\textsc{unsat} & \textcolor{darkgray}{2.42} & \textcolor{darkgray}{7.02} & \textcolor{darkgray}{1.58} & \textcolor{darkgray}{3.15} & \textcolor{darkgray}{0.76} & - & - \\
2023 Acasxu & 6 & ~\textsc{unsat} & \textcolor{darkgray}{2.29} & \textcolor{darkgray}{7.02} & \textcolor{darkgray}{1.38} & \textcolor{darkgray}{3.18} & \textcolor{darkgray}{0.50} & - & - \\
2023 Acasxu & 7 & ~\textsc{unsat} & \textcolor{darkgray}{2.21} & \textcolor{darkgray}{7.02} & \textcolor{darkgray}{1.14} & \textcolor{darkgray}{3.14} & \textcolor{darkgray}{0.58} & - & - \\
\bottomrule
\end{longtable}
}
\end{center}


\end{document}